\documentclass[10pt,twocolumn,letterpaper]{article}

 \usepackage{bm}   
 \usepackage{longtable}
 \usepackage{supertabular}
 \usepackage{mathtools}
 \usepackage{algorithm}
 \usepackage{algpseudocode}
 \usepackage{mathtools}
\usepackage{subcaption}
\usepackage{multirow}
\usepackage{utfsym}
\usepackage{array, booktabs, makecell}
\usepackage[version=4]{mhchem}
\usepackage{siunitx}
\usepackage[pagenumbers]{cvpr} 
\usepackage{caption}
\usepackage{bbm}
\usepackage[spacing=true,factor=1100,stretch=10,shrink=10]{microtype}

\definecolor{cvprblue}{rgb}{0.21,0.49,0.74}
\usepackage[pagebackref,breaklinks,colorlinks,citecolor=cvprblue]
{hyperref}

\def\paperID{13037} 
\def\confName{CVPR}
\def\confYear{2024}

\title{Large-scale Reinforcement Learning for Diffusion Models}

\author{Yinan Zhang\\
Pinterest ATG\\
\and
Eric Tzeng \\
Pinterest ATG\\
\and
Yilun Du \\
MIT CSAIL \\ 
\and 
Dmitry Kislyuk \\
Pinterest ATG
}

\renewcommand{\paragraph}[1]{\noindent\textbf{#1.}}
\begin{document}
\maketitle

\begin{abstract}
 
 Text-to-image diffusion models 
 are a class of deep generative models that 
 have demonstrated an impressive capacity for high-quality image generation. However, these models are susceptible to implicit biases that arise from web-scale text-image training pairs and may inaccurately model aspects of images we care about. This can result in suboptimal samples, model bias, and images that do not align with human ethics and preferences. In this paper, we present an effective scalable algorithm to improve diffusion models using Reinforcement Learning (RL) across a diverse set of reward functions, such as human preference, compositionality, and fairness over millions of images. We illustrate how our approach substantially outperforms existing methods for aligning diffusion models with human preferences. We further illustrate how this substantially improves pretrained Stable Diffusion (SD) models, generating samples that are preferred by humans 80.3\% of the time over those from the base SD model while simultaneously improving both the composition and diversity of generated samples. The project’s website can be found at \href{https://pinterest.github.io/atg-research/rl-diffusion/}{https://pinterest.github.io/atg-research/rl-diffusion/}.
\end{abstract}    
\section{Introduction}
\label{sec:intro}
 Diffusion probabilistic models \cite{sohldickstein2015deep, ho2020denoising, rombach2022highresolution} have revolutionized generative modeling, particularly for producing creative and photorealistic imagery when combined with pre-trained text encoders \cite{radford2021learning, raffel2023exploring}. 
 However, the resulting image quality is highly dependent on the distribution of the pre-training dataset, which typically consists of web-scale text-image pairs.
 Although pre-training on massive weakly supervised tasks of this form is effective in exposing the text-to-image model to a wide range of prompts, downstream applications often observe weaknesses around the following properties:
 \begin{itemize}
     \item \textbf{Fidelity and controllability} \cite{cho2023dalleval, gokhale2023benchmarking, huang2023t2icompbench}: failing to accurately depict the semantics of the text prompts (e.g. incorrect composition and relationships between objects)
     \item \textbf{Human aesthetic mismatch} \cite{xu2023imagereward, wu2023human}: producing outputs that humans do not perceive to be aesthetically pleasing 
     \item \textbf{Bias and stereotypes} \cite{smith2023balancing, Bianchi_2023,luccioni2023stable}: presenting or exaggerating societal bias and stereotypes
     
 \end{itemize}

To address these challenges, several works have explored classic fine-tuning techniques for pre-trained diffusion models with curated data, either to improve the aesthetic quality of the model outputs with human-selected high-quality images \cite{dai2023emu}, or to eliminate existing biases in the model with synthetic dataset augmentation \cite{esposito2023mitigating}. Another approach, which bypasses the labor-intensive dataset curation, involves intervention in the sampling process to achieve controllability, by utilizing auxiliary input \cite{li2023gligen, lian2023llmgrounded, feng2023layoutgpt} or refining the intermediate representations \cite{chefer2023attendandexcite,feng2023trainingfree, chen2023trainingfree}. However, this form of inference-time guidance results in an increase in the sampling time without improving the inherent capability of the model.
A recent direction, motivated by the success of reinforcement learning from human feedback (RLHF) in the language domain \cite{ouyang2022training, nakano2022webgpt, bai2022training}, proposes \cite{xu2023imagereward, clark2023directly} fine-tuning diffusion models through full-sample gradient backpropogation on human preference reward models, though these approaches are memory intensive and only work for differentiable reward functions. 
Finally, RL-based optimization \cite{black2023training, fan2023dpok} has enabled fine-tuning with arbitrary objective functions, but these methods have so far been limited in scope by focusing on a small set of prompts in a narrow domain, and lack the scale to improve model performance generally. 



In this paper, we propose a generic RL-based framework for fine-tuning diffusion models, which works at scale across millions of prompts and with an arbitrary combination of objective functions. Our contributions are as follows:


\begin{itemize}
    \item   We present an effective large-scale RL training algorithm for diffusion models which allows training over millions of prompts across a diverse set of tasks.
    \item  We propose a distribution-based reward function for RL fine-tuning to improve the output diversity.
    \item  We demonstrate how to perform effective \textit{multi-objective} RL-training and illustrate how we can improve a base model across all objectives, which can include human aesthetic preference, fairness, and object composition.
    \item We conduct extensive experiments and analysis studies comparing our approach with existing reward optimization methods across a suite of tasks.
\end{itemize}

\begin{figure*}[t]
  \centering
   \includegraphics[width=0.9\linewidth]{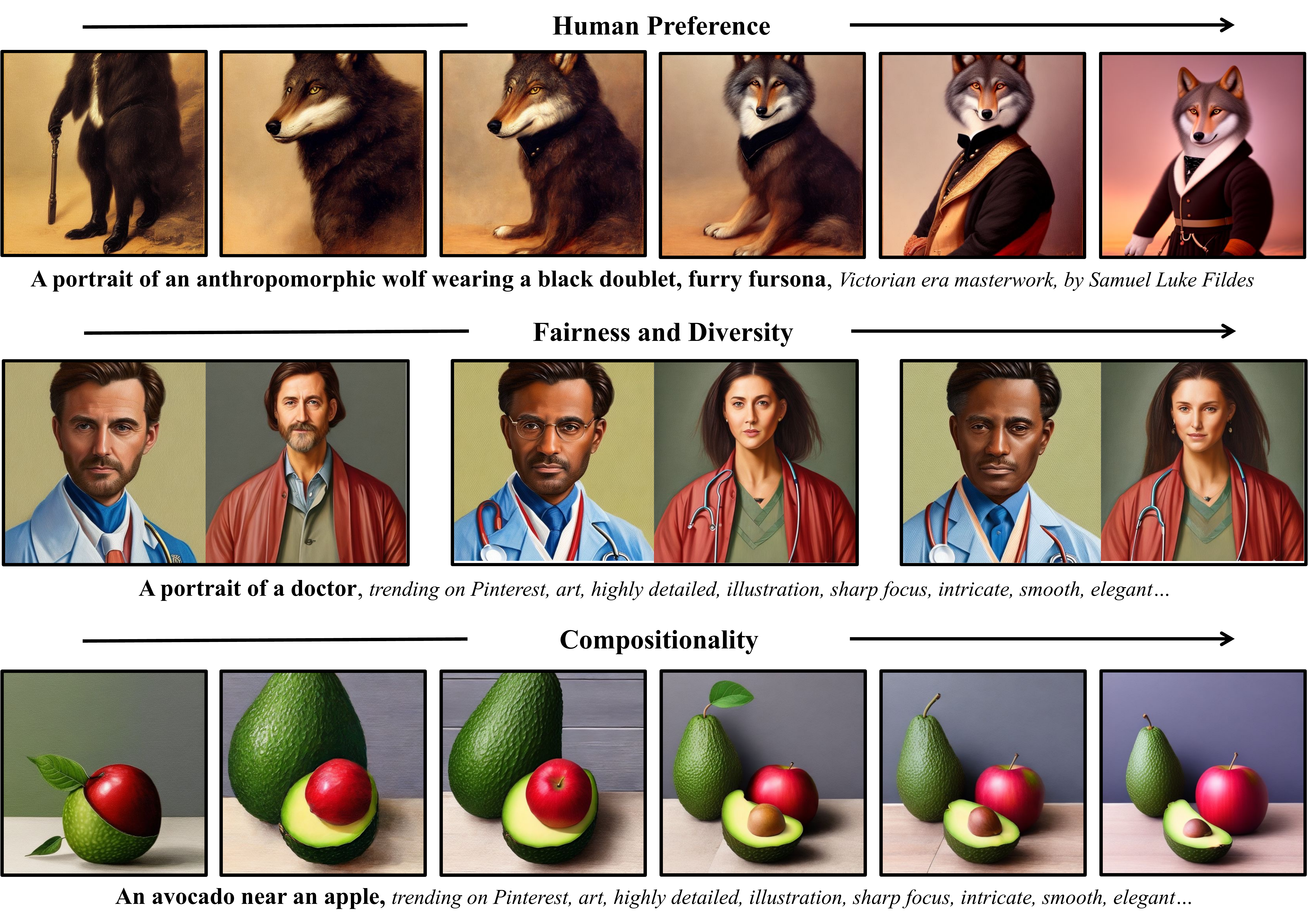}
    \vspace{-10pt}
   \caption{\textbf{Sample Evolution over Reinforcement Training.} We perform multi-task RL on text-to-image diffusion models, improving the model's compositional capacity and alignment with human preference while mitigating its bias and stereotypes. Here we show the progression of samples over training across each objective, with the leftmost columns showing results from the base SDv2 model.}
   \vspace{-10pt}
   \label{fig:progression_3training}
\end{figure*}



\section{Related Work}
\label{sec:related_work}

\paragraph{Reward Fine-tuning for Diffusion Models} Existing reward fine-tuning methods for diffusion models can be classified into three categories: either supervised with reward-weighted data~\cite{lee2023aligning,wu2023human, dong2023raft}, optimized through gradient-backpropogation on the reward function~\cite{xu2023imagereward,clark2023directly} or through reinforcement learning~\cite{black2023training,fan2023dpok}. Our work builds on work training diffusion models with reinforcement learning, but while past work has focused on simple settings (DPOK uses a training set of 1 prompt per model, and DDPO using simple set of 45 common animals and 3 activities), we illustrate how we can use reinforcement learning training across the scale of millions of prompts and different objectives. 

\paragraph{Compositional Text-to-image Generation} Despite their remarkable capacity, current state-of-the-art text-to-image models still struggle to generate images that faithfully align with the semantics of the text prompts due to their limited compositional capabilities~\cite{cho2023dalleval, gokhale2023benchmarking, huang2023t2icompbench}. Existing work addresses this by either modifying the inference procedure ~\cite{liu2022compositional,chefer2023attendandexcite, feng2023trainingfree, feng2023trainingfree, du2023reduce} or by using auxiliary conditioning inputs such as bounding boxes \cite{chen2023trainingfree, li2023gligen} or spatial layouts \cite{wu2023harnessing,  lian2023llmgrounded, feng2023layoutgpt}. Our method instead focus on improving the fidelity of existing SD models without using additional layout guidance. 

\paragraph{Inclusive Text-to-Image Generation}
Text-to-image generative models perpetuate and even amplify the societal biases present in the massive pretraining datasets of uncurated image-text pairs \cite{cho2023dalleval, zhang2023auditing, changpinyo2021conceptual,CheongManuscript-CHEIGA-2}. Existing work addresses this by either using balanced synthetic data  \cite{smith2023balancing}, with textual guidance during inference \cite{friedrich2023fair} or with reference images of a particular attribute \cite{zhang2023inclusive}. Different from prior work, our method does not require synthetic data collection or inference-time intervention.

\vspace{-7pt}
\section{Method}
\vspace{-3pt}

\label{sec:method}
In this section, we describe our approach for applying large-scale RL training to diffusion models. Our goal is to fine-tune the parameters $\bm{\theta}$ of an existing diffusion model to maximize the reward signal $r$ of the generated images from the sampling process:
\begin{equation}
    J(\bm{\theta}) = \mathbb{E}_{\bm{c}\sim p(\bm{c}), \bm{x}_0\sim p_{\bm\theta}(\bm{x}_0|\bm{c})}[r(\bm{x}_0,\bm{c})], 
\end{equation}
where $p(\bm{c})$ is the context distribution, $p_{\bm\theta}(\bm{x}_0|\bm{c})$ is the sample distribution, and $r(\bm{x}_0,\bm{c})$ is the reward function  that is applied to the final sample image.

\subsection{Policy Gradient with Multi-step MDP} Following Black~\etal~\cite{black2023training}, we reframe the iterative denoising procedure of diffusion models as a multi-step Markov decision process (MDP), where the policy, action, state and reward at each timestep $t$ are defined as follows:
\begin{align}
\pi(\bm{a}_t|\bm{s}_t) &\triangleq p_{\bm\theta}(\bm{x}_{t-1}|\bm{x}_t,\bm{c}) \\
\bm{a}_t &\triangleq \bm{x}_{t-1}   \\
\bm{s}_t &\triangleq (\bm{c}, t, \bm{x}_t) \\
R(\bm{s}_t, \bm{a}_t) &\triangleq 
\begin{cases}
      r(\bm{x}_0, \bm{c}) & \text{if $t=0$}\\
      0 & \text{otherwise}
    \end{cases} 
\end{align}
We treat the reverse sampling process $p_{\bm\theta}(\bm{x}_{t-1}|\bm{x}_t,\bm{c})$ of the diffusion model as the policy. Starting from a sampled initial state $\bm{x}_T$, the policy's action at any timestep $t$ is the update that produces the sample for the next timestep $\bm{x}_{t-1}$. The reward is defined as $r(\bm{x}_0,\bm{c})$ at the final timestep, and $0$ otherwise. 

The policy gradient estimates can be made using the likelihood ratio method (also known as REINFORCE) \cite{Williams2004SimpleSG, mohamed2020monte}: 
\begin{equation}
    \nabla_{\bm{\theta}}J = \mathbb{E} \left[ r(\bm{x}_0, \bm{c})\sum_{t=0}^{T}\nabla_{\bm{\theta}} \log p_{\bm\theta}(\bm{x}_{t-1}|\bm{x}_t,\bm{c})\right].
\end{equation}
We also apply importance sampling to enable collecting samples from the old policy for improved training efficiency, and incorporate a clipped trust region to ensure that the new policy does not deviate too much from the old policy~\cite{schulman2017proximal}. The final clipped surrogate objective function can be written as:
\vspace{-5pt}
\begin{equation}
\label{eq:clip_obj}
 \resizebox{0.9\hsize}{!}{$
J(\bm{\theta}) = \mathbb{E} \left[ \sum_{t=0}^{T} \min \left[ w({\bm \theta}, {\bm \theta_\text{old}})
 \hat{A}(\bm{x}_0, \bm{c}), g(\epsilon,\hat{A}(\bm{x}_0,\bm{c})) \right] \right] $}
\end{equation}
where
\vspace{-5pt}
\begin{gather*}
w({\bm \theta}, {\bm \theta}_\text{old}) = \frac{p_{\bm\theta}(\bm{x}_{t-1}|\bm{x}_t,\bm{c})} {p_{\bm\theta_{\text{old}}}(\bm{x}_{t-1}|\bm{x}_t,\bm{c})},\\
g(\epsilon,A) =
\begin{cases}
      (1+\epsilon)A & \text{if $A\geq0$}\\
      (1-\epsilon)A & \text{if $A<0$}
    \end{cases}.
\end{gather*}
Here $\epsilon$ is the hyper-parameter that determines the clip interval, and $\hat{A}(\bm{x}_0, \bm{c})$ is the estimated \textit{advantage} for the samples.
To further prevent over-optimization of the reward function, we also incorporate the original diffusion model objective as part of the loss function.
Our full training objective is thus
\begin{equation}
    L(\bm{\theta}) = J(\bm{\theta}) + \beta L_\text{pre}(\bm{\theta}),
\end{equation}
where
\vspace{-3pt}
\begin{equation}
\label{eq:pretraining_loss}
L_\text{pre}(\bm{\theta}) = \mathbb{E}_{\varepsilon(x), \epsilon \sim \mathcal{N}(0,1),t} \left[ \lVert \epsilon - \epsilon_{\theta}(z_t, t) \rVert ^2 _2 \right].
\end{equation}
\looseness=-1
One additional detail is that reward values are typically normalized to zero mean and unit variance during gradient updates to increase training stability. In policy-based RL, a general approach is to subtract a baseline state value function from the reward to obtain the \textit{advantage function}~\cite{NIPS1999_464d828b}
\begin{equation}
\hat{A}(\bm{x}_0, \bm{c}) = \frac{r(\bm{x}_0, \bm{c})-\mu_r}{\sqrt{\sigma_{r}^2+\epsilon}}.
\end{equation} 
In the original implementation of DDPO, Black~\etal normalize the rewards on a per-context basis by keeping track of a running mean and standard deviation for each prompt independently~\cite{black2023training}. However, this approach remains impractical if the training set size is unbounded or unfixed.

In contrast to the limited size of their training prompts (up to $398$ only), our large-scale fine-tuning experiments involve millions of training prompts. We instead normalize the rewards on a per-batch basis using the mean and variance of each training batch.

\subsection{Distribution-based Reward Functions}
\label{sec:3.3}



\begin{figure*}[t]
\centering


  

\begin{subfigure}[t]{\textwidth}
\centering
   \includegraphics[width=\linewidth]{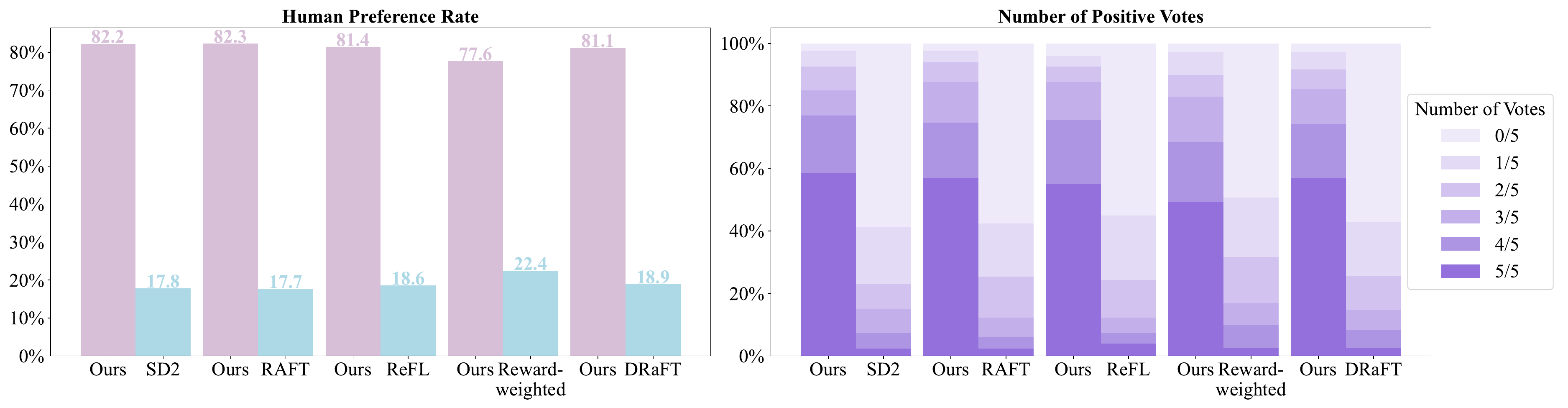}
  \caption{Human Eval Results on DiffusionDB~\cite{wangDiffusionDBLargescalePrompt2022} dataset}
  \label{fig:dbprompts_human_eval}
\end{subfigure}

\begin{subfigure}[t]{\textwidth}
\centering
   \includegraphics[width=\linewidth]{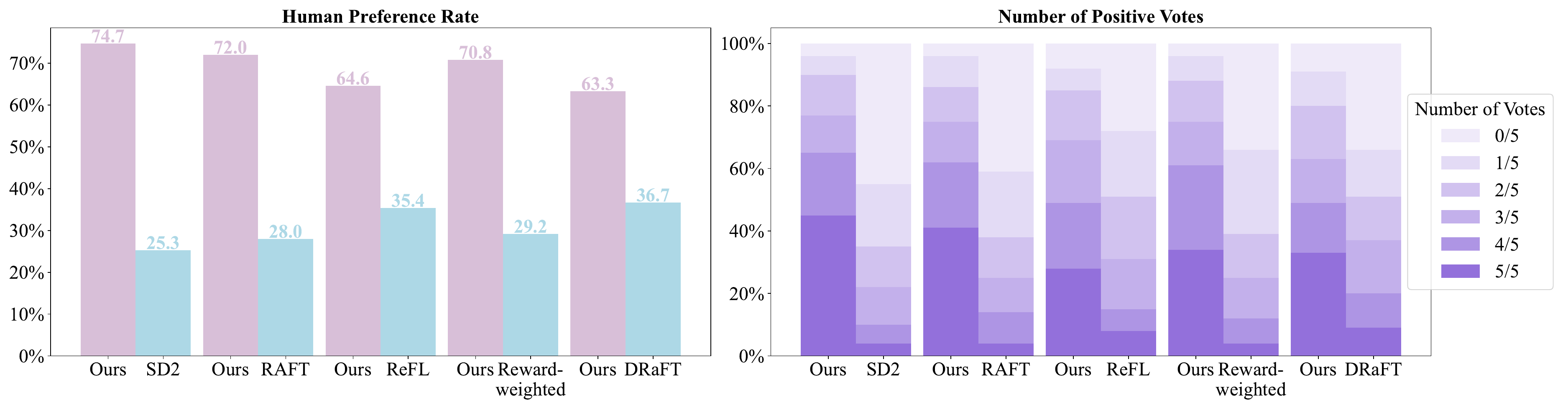}
  \caption{Human Eval Results on PartiPrompts~\cite{yu2022scaling} dataset}
  \label{fig:parti_human_eval_nov}
\end{subfigure}

\vspace{-5pt}
\caption{\textbf{Human Preference Evaluation of Generations.} Human evaluation results on 400 text prompts (300 randomly sampled from DiffusionDB dataset and 100 randomly sampled from PartiPrompts dataset). 
We perform head-to-head comparisons between images generated by our model and each of the baseline models, using the same text prompt and random seed for each generation.
Then, human raters indicate which one is better in terms of image quality and image-text alignment.
Each query is evaluated by 5 independent human raters, and we report each model's preference rate based on the number of positive votes it received.}
\label{fig:human_eval}
\vspace{-10pt}
\end{figure*}

In the previously outlined formulation of the diffusion MDP, each generation is considered independently, and thus rewards incurred by generated samples are independent of each other.
This formulation is a natural fit for reward functions that only care about the contents of a single image, such as image quality or text-image alignment.
However, sometimes what we care about is not the contents of any particular image, but instead the output distribution of the diffusion model as a whole.
For example, if our goal is to ensure our model generates diverse outputs, considering a single generation in isolation is insufficient---we must consider the set of all outputs in order to understand these distributional properties of our model.

To this end, we also investigate the use of distribution-level reward functions for reinforcement learning with diffusion models.
However, it is intractable to construct the true generative distribution. Thus, we instead approximate the reward by computing it using empirical samples across minibatches during the reinforcement learning process.
During training, the attained reward is computed on each minibatch, and the minibatch reward is then backpropagated across the samples to perform model updates.
In Section~\ref{sec:4.2diversity} we validate this approach by learning via a distribution-level reward function that optimizes for fairness and diversity in generated samples.

\subsection{Multi-task Joint Training}
We also perform multi-task joint training to optimize a single model for a diverse set of objectives simultaneously. As detailed in the next section, we incorporate the reward functions from human preference, skintone diversity, object composition and perform joint-optimization all at once. Since each task involves a different distribution of training prompts, in every training iteration, we sample multiple prompts from all the tasks and run the sampling process independently. Each reward model is applied to the corresponding sample image with the prompt. Then the gradient step from equation~\ref{eq:clip_obj} is executed for each task sequentially. We outline the training framework in Algorithm~\ref{alg:multi-reward} with hyper-parameters available in Appendix \ref{supp:hyperparameters}.

\begin{algorithm}[t]
\caption{Multi-reward diffusion policy optimization}
\label{alg:multi-reward}
\begin{algorithmic}
\renewcommand{\algorithmicrequire}{\textbf{Input:}}
\renewcommand{\algorithmicensure}{\textbf{Output:}}

\algnewcommand\algorithmicforeach{\textbf{for each}}
\algdef{S}[FOR]{ForEach}[1]{\algorithmicforeach\ #1\ \algorithmicdo}
\Require{A set of reward models and the training prompt distribution $S=\{(r_i,p_i({\bm{c}}))\}$, pretrained diffusion model $p_{pre}$, current diffusion model $p_{\bm{\theta}}$}, 
pretraining dataset $D^{pre}$
\State Initialize $p_{\bm{\theta}} = p_{pre}$
\While {$\theta$ not converged}
\State $p_{\bm\theta_{\text{old}}} = p_{\bm\theta}$
\ForEach {\text{training task } $(r,p({\bm{c}})) \in \mathcal S $}
\State Sample a prompt $\bm{c} \sim p({\bm{c}})$
\State Sample generated images $\bm{x}_{0:T}\sim p_{\bm\theta}(\bm{x}_{0:T}|\bm{c})$
\State Sample training timesteps $\bm{t}$
\For{each selected timestep $t$}
        \State Take gradient step $\nabla_{\bm{\theta}} J({\bm \theta})$ (Eq.~\ref{eq:clip_obj})
      \EndFor
\EndFor
\State Sample a pretraining data pair $(txt, img) \in D^{pre}$
\State Take gradient step $\nabla_{\bm{\theta}}L_{pre}(\bm{\theta})$ (Eq.~\ref{eq:pretraining_loss})
\EndWhile
\Ensure{Fine-tuned diffusion model $p_{\bm\theta}$}
\end{algorithmic}
\end{algorithm}

\section{Reward Functions and Experiments}
\label{sec:rewards}

To validate our method across a wide variety of settings, we perform experiments on three separate reward functions: human preference, image composition, and diversity and fairness.
We begin with an introduction of the different reward functions we applied our method to.

To optimize diffusion models to adhere to human preferences, we use an open-source reward model, ImageReward (IR), trained on a large number of human preference pairs~\cite{xu2023imagereward}.
ImageReward takes a pair consisting of a text caption and a generated sample, then outputs a human preference score, which is then used as the reward during training:
\begin{equation}
    \label{eq:reward-humanpreference}
    r(x_0, c) = \text{IR}(x_0, c).
\end{equation}
Our results with this human preference reward function are detailed in Section~\ref{sec:4.1humanpreference}.

In order to encourage fairness and diversity across the samples generated by our model, following previous work~\cite{choi2020fair, teo2021measuring, chuang2023debiasing}, we leverage \textit{statistical parity}, a metric commonly adopted for measuring biases in models, as a distribution-level reward function for our fine-tuning experiments.
Given the generated distribution $\hat{P}$ and a classifier $h: x \xrightarrow{} \mathcal{A}$ that identifies a spurious attribute, we measure the L2 norm between the empirical and uniform distributions:
\begin{equation}
\label{eqn:statisticalparity}
\sqrt{\sum_{a\in\mathcal{A}}(\mathbb{E}_{x\sim \hat{P}}\left[ \mathbbm{1}_{h(x)=a} \right] - 1/|\mathcal{A}|)^2}
\end{equation}
%
The reward attained by the model is then simply the negation of the statistical parity, so as to encourage the model to produce diverse samples.
As explained in Section~\ref{sec:3.3}, it is intractable to compute the reward over the full output distribution of the model, so we compute the reward over individual minibatches.
We present the results for this experiment in Section~\ref{sec:4.2diversity}.

To improve the compositional skills of diffusion models, we devise a new reward function that uses an auxiliary object detector.
We construct a set of training prompts, each containing multiple different objects, and use an object detection model on the image to predict the confidence score for each object class. The reward score is then defined as the average confidence score of all the objects:
\begin{equation}
    \label{eq:reward-compositionality}
    r(x_0, c) = \frac{1}{|o|}\sum_{o \in c}d(o, x_0),
\end{equation}
where $d(o, x_0)$ is the detection confidence score for the object class $o$ given input image $x_0$.
Our results on compositionality are detailed in Section~\ref{sec:4.3compositionality}.

Finally, we also experiment with jointly optimizing over all three previously described reward functions, to train a model that satisfies all three criteria simultaneously. We present the results of our joint  optimization in Section~\ref{sec:4.4multireward}. For all our fine-tuning experiments, we use SDv2~\cite{rombach2022highresolution} as our base model. The output resolution is $512$x$512$, which we consider as a good tradeoff between compute efficiency and image quality.


\subsection{Learning from human preference}
\label{sec:4.1humanpreference}

To fine-tune a diffusion model with human preferences, we use ImageReward~\cite{xu2023imagereward}, which was trained on large-scale human assessments of text-image
pairs. In total, the authors collected 137k pairs of expert judgments on images generated from real-world user prompts from the DiffusionDB dataset~\cite{wangDiffusionDBLargescalePrompt2022}. Compared to other existing metrics such as CLIP~\cite{radford2021learning}, BLIP~\cite{li2022blip}, or Aesthetic score~\cite{schuhmann2022laion5b}, ImageReward is better aligned with human judgments, making it better suited as a reward function. 

We use a training set of 1.5 million unique real user prompts from DiffusionDB, among which 2,000 prompts were split for testing. We use 128 A100 GPUs (80GB) for all experiments, including the baselines. Experimental details, hyperparameters,
and additional results are provided in Appendix \ref{supp:hyperparameters}.

\looseness=-1
\paragraph{Baseline Comparison} Prior reward fine-tuning methods for diffusion models mainly fall under three categories: reward-based loss reweighting~\cite{lee2023aligning}, dataset augmentation~\cite{dong2023raft}, and backpropagation through the reward model~\cite{xu2023imagereward, clark2023directly}. We compare against a variety of baseline methods, including ReFL~\cite{xu2023imagereward}, RAFT~\cite{dong2023raft}, DRaFT~\cite{clark2023directly} and Reward-weighted~\cite{lee2023aligning}, covering the three different methodologies. We reimplement all methods and fine-tune them on SDv2 using the same training set of 1.5M prompts until convergence. 

\begin{figure*}[!tb]
  \centering
   \includegraphics[width=\linewidth]{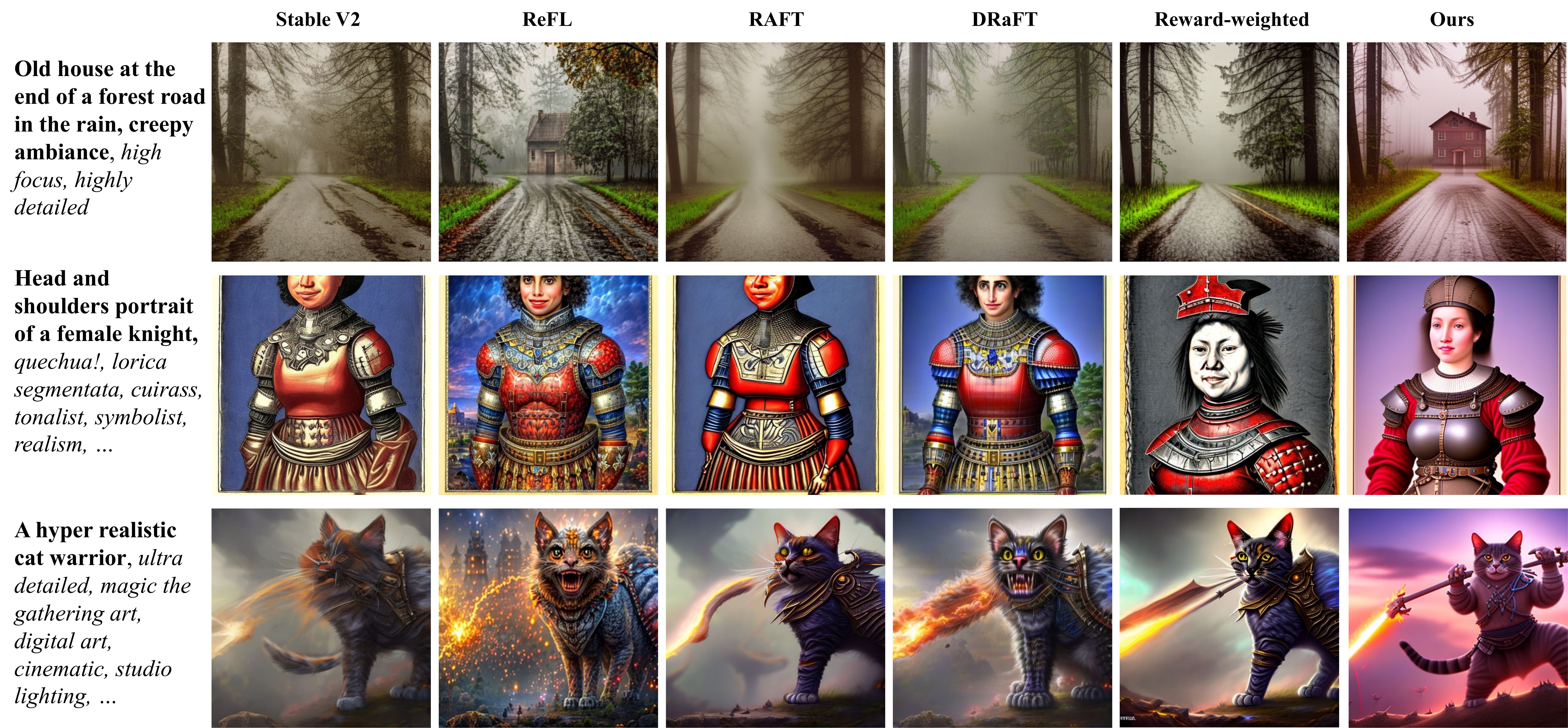}
   \caption{\textbf{Qualitative comparison of our approach and other reward fine-tuning methods on real-user prompts}. All images are generated using the same random seeds.}
   \label{fig:baseline_vis_6x6}
   \vspace{-5pt}
\end{figure*}

\def\thickhline{\noalign{\hrule height.8pt}}
      
\begin{table}[t]

\centering 
\resizebox{0.45\textwidth}{!}{%
  \begin{tabular}{l c c c c}
    \toprule
    \multirow{2}{*}{\textbf{Model}} &
      \multicolumn{2}{c}{\textbf{DiffusionDB}} &
      \multicolumn{2}{c}{\textbf{PartiPrompts}} \\
      \cmidrule(lr){2-3} \cmidrule(lr){4-5} %

    & IR* & Aesthetic  & IR* & Aesthetic \\
    \midrule
    Stable v1.5 & 0.082  & 5.907 & 0.256  & 5.373 \\
    Stable v2 & 0.170 & 5.783  & 0.414 & 5.269 \\
    ReFL  & \textbf{1.290} & 5.845 &\textbf{0.832} & 5.402  \\
    RAFT & 0.338  & 5.881 & 0.504 & 5.413 \\
    DRaFT & 0.818 & 5.645 & 0.632 & 5.279  \\
    Reward-weighted & 0.438 & 5.821 & 0.624 & 5.363  \\ \midrule
    Ours & 0.845 & \textbf{5.918} & 0.731 & \textbf{5.477} \\
    \bottomrule
  \end{tabular}%
}
\vspace{-5pt}
\caption{\textbf{Quantitative Results.} ImageReward scores and Aesthetic scores from the original SDv2 model, baseline methods, and our model. We report the average ImageReward and Aesthetic scores for samples generated using prompts from both the DiffusionDB~\cite{wangDiffusionDBLargescalePrompt2022} dataset and the PartiPrompts~\cite{yu2022scaling} dataset.}

\label{table:imagereward_quantiative} 
\end{table}

\begin{figure*}[t]
  \centering
   \includegraphics[width=0.8\linewidth]{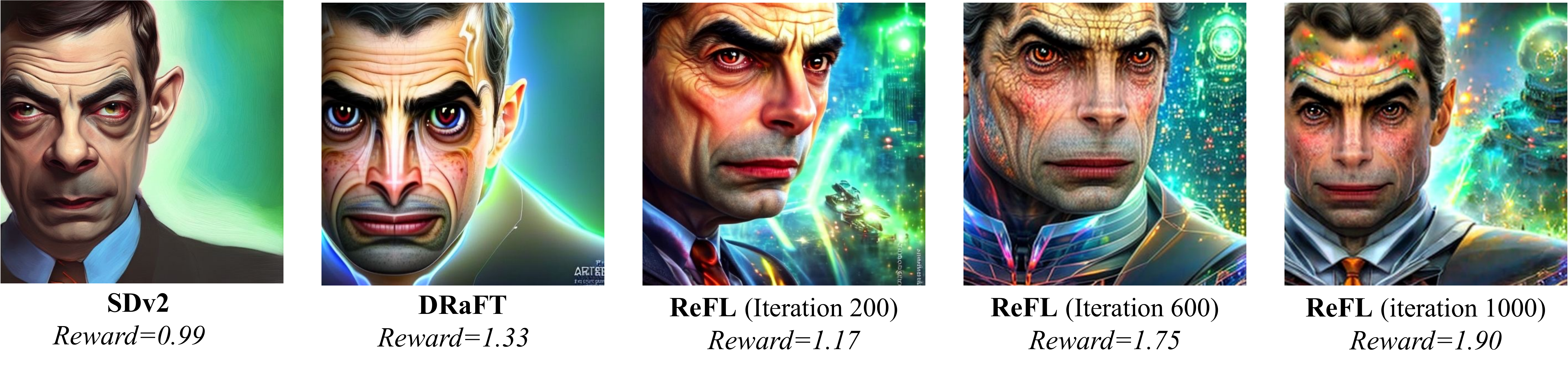}
   \vspace{-10pt}
   \caption{\textbf{Reward Hacking.} Finetuning methods such as  DRaFT and different iterations of ReFL fine-tuned models often over-optimize reward functions and generate over-detailed images with high-frequency noise.}
   \label{fig:reward_hacking}
   \vspace{-15pt}
\end{figure*}

We show the qualitative and quantitative results of all baseline methods in Figure \ref{fig:baseline_vis_6x6} and Table  \ref{table:imagereward_quantiative}. 
We also provide training curves in Appendix \ref{supp:training_curve} and note that, except for RAFT which diverged, all online-learning methods exhibit steadily increasing sample rewards during training, eventually saturating at some maximum level, at which point we consider the models converged. In contrast to the common belief that RL training is inefficient and slow to converge,  our approach converges in as few as $\sim$1,000 steps, compared to DRaFT, the gradient-based reward optimization approach which takes $\sim$4,000 steps to converge while only being able to optimize for differentiable rewards. 
We provide a comprehensive comparison of all the reward optimization methods in Table~\ref{table:baseline_details}. 

\def\thickhline{\noalign{\hrule height.8pt}}
\begin{table}[!ht]
\newcommand{\cmark}{\usym{2713}}%
\newcommand{\xmark}{\usym{2717}}%

\centering 
\vspace{-5pt}
\resizebox{\linewidth}{!}{%
  \begin{tabular}{l l c c }
    \toprule 
     & Training & Generalizable to  & Human  Preference \\
     Model & Time & All Rewards & Rank \\
    \midrule
    RAFT & 5.5h (diverged)  & \cmark & 5 \\
    ReFL & 6.9h  & \xmark & 4 \\
    DRaFT & 8.4h  & \xmark  & 3 \\
    Reward-weighted & 33.8h$^*$  & \cmark & 2 \\ 
    Ours & 12.1h  & \cmark  & 1  \\
    \bottomrule
  \end{tabular}
  }
\vspace{-5pt}
\caption{\textbf{Performance Comparison.} Comparison of different reward optimization methods. Training time indicates the time for each method to converge. $^*$For Reward-weighted, training time includes constructing the training dataset from the base model.}
\label{table:baseline_details} 
\vspace{-10pt}
\end{table}

For RAFT, we found the model diverges as the number of training iterations increases, similar to the finding from Xu \etal~\cite{xu2023imagereward}. Since RAFT uses the model-generated images with the highest rewards for fine-tuning the model, it is constrained by the diversity of the latest model's generation and thus prone to overfitting. The reward-weighted method uses a similar idea of augmenting the training data using model-generated images and weighting the training loss by the reward values, but all the images are generated from the original model (in contrast to RAFT's online generation using the latest model) and thus is less prone to overfitting.

\begin{figure*}[t]
  \centering
   \includegraphics[width=\linewidth]{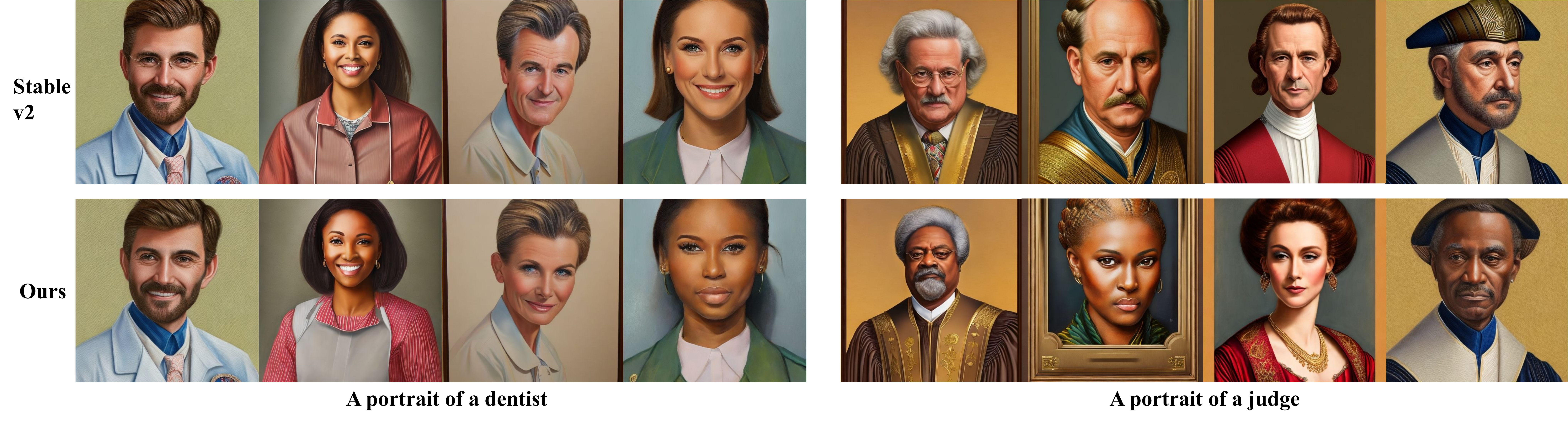}
   \vspace{-22pt}
   \caption{\textbf{Skintone Diversity Visualization.} Qualitative comparison of SDv2 and our model fine tuned for skintone diversity reward. All images are generated using the same random seeds.}
   \label{fig:sd2_ours_skintone}
   \vspace{-15pt}
\end{figure*}

\paragraph{Evaluating generalization} Next, we evaluate our trained model's ability to generalize to an out-of-domain test set, PartiPrompts \cite{yu2022scaling}. PartiPrompts is a comprehensive benchmark for text-to-image models, with over 1,600 challenging prompts across a variety of categories.
We report the ImageReward and Aesthetic scores in Table~\ref{table:imagereward_quantiative}, along with human evaluation results in Figure~\ref{fig:human_eval}. When compared against each baseline model, our approach achieves the highest Aesthetic score and human preference rate on both sets.

\looseness=-1
We also achieve the second highest result on ImageReward, but note that this metric alone is not a robust indicator of performance, since the model was directly trained against it.
Reward hacking is a commonly observed phenomenon in which models optimizing for a single reward function often overoptimize for this single metric at the cost of overall performance.
We believe the high ImageReward scores achieved by ReFL are a result of this, and show example generations in Figure~\ref{fig:reward_hacking}.
The reward hacking problem of ReFL was observed by Clark \etal~\cite{clark2023directly} in their DRaFT experiments as well, where their fine-tuned model optimizing for Aesthetic score collapses to generate very similar, high-reward images.
We hypothesize that gradient-based optimization methods (i.e. ReFL and DRaFT) are more prone to reward hacking due to their direct access to the gradients of the reward model. 
In contrast, our wins on human preference rate indicate that our method is more robust to these effects.


\subsection{Optimizing Fairness and Diversity}
\label{sec:4.2diversity}

The training of diffusion models is highly data-driven, relying on billion-sized datasets that are randomly scraped from internet. As a result, the trained models may contain significant social bias and stereotypes. For example, 
it has been observed that text-to-image diffusion models commonly exhibit a tendency to generate humans with lighter skintones~\cite{naik2023social, CheongManuscript-CHEIGA-2}.  We aim to mitigate this bias by explicitly guiding the model using a skintone diversity reward.

For fine-tuning, we collect a dataset of 240M human images from Pinterest and run BLIP~\cite{li2022blip} to generate captions for each image. Only the text prompts are used during training, and the reward calculation is based on the generated samples. We further filter out the captions containing terms relating to ethnicity and race (e.g. African, Asian, Indian) to ensure that the training prompts are race agnostic. In each training iteration, we load 128 prompts and generate a minibatch of 16 images for each prompt, then run a pre-trained skintone classifier on the generated samples and calculate the statistical parity for each minibatch according to Equation~\ref{eqn:statisticalparity}. Since the classifier has 4 skintone categories ranging from dark to light, the optimal reward is achieved when the output distribution is 
entirely uniform (i.e. 4 samples in each skintone bucket).

\def\thickhline{\noalign{\hrule height.8pt}}
      
\begin{table}[!ht]
\small
\centering 
\vspace{-5pt}
  \begin{tabular}{l c c}
    \toprule
    \multirow{2}{*}{\textbf{Model}} &
      \multicolumn{2}{c}{\textbf{Statistical Parity ($\downarrow$)}}  \\
      \cmidrule(lr){2-3} %
    & Occupation & HRS-Bench \\
    \midrule
    Stable v1.5 &  0.575 & 0.578  \\
    Stable v2 & 0.556 &  0.576 \\

    RAFT & 0.464 & 0.527 \\
    Reward-weighted & 0.562  & 0.527  \\
    Ours & \textbf{0.453} & \textbf{0.498} \\
    \bottomrule
  \end{tabular}%
\vspace{-5pt}
\caption{\textbf{Fairness and Equity Evaluation.} Statistical Parity scores on out-of-domain test sets.} 
\vspace{-8pt}
\label{table:skintone} 
\end{table}

We show our qualitative results in Figures \ref{fig:progression_3training} and \ref{fig:sd2_ours_skintone} and quantitative results in Table \ref{table:skintone}. We construct two test sets: a set of 100 randomly sampled occupations, for which we add the prefix ``a portrait of'' to produce the final prompts (e.g.``a portrait of a police officer''), and another set of 200 prompts from HRSBench~\cite{bakr2023hrsbench}, which are descriptions of people with random objects. We note that both are out-of-domain test sets, as their distribution is different from that of the BLIP-generated training prompts.

Our fine-tuned model greatly reduces the skintone bias embedded in the pretrained SDv2 model, especially for occupations with more social stereotypes or biases inherent in the pretraining dataset. For example, in Figure~\ref{fig:sd2_ours_skintone}, we show that the pretrained SDv2 model is biased towards light skintone for portraits of dentists and judges, whereas our finetuned model generates a much more balanced distribution.



\subsection{Optimizing Compositionality}

\label{sec:4.3compositionality}
\begin{figure*}[!tb]
  \centering
   \includegraphics[width=0.95\linewidth]{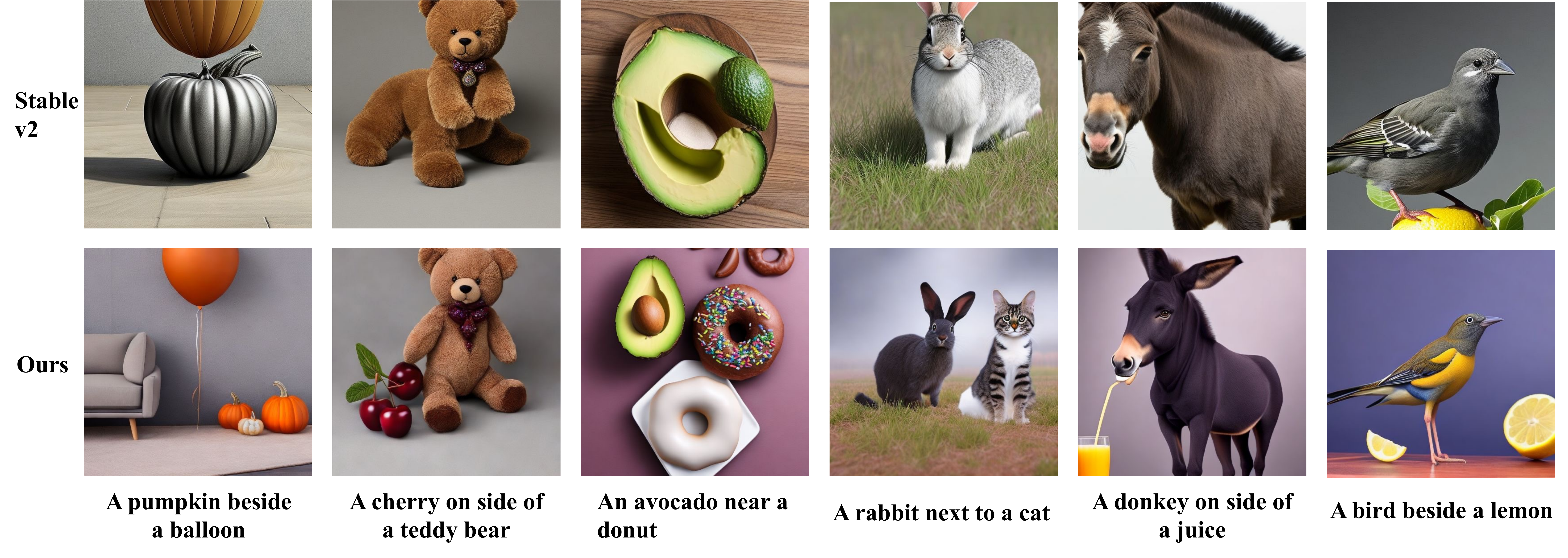}
   \vspace{-10pt}
   \caption{\textbf{Compositional Visualization.} Qualitative comparison of SDv2 and our model fine-tuned for compositionality reward. All images are generated using the same random seeds.}
   \label{fig:sd2_ours_compositionality}
   \vspace{-5pt}
\end{figure*}

While diffusion models are able to generate diverse images, they often fail to accurately generate different compositions of objects in a scene~\cite{liu2022compositional, gokhale2023benchmarking, huang2023t2icompbench, cho2023dalleval}. We further explore using our RL framework in ensuring compositionality with diffusion models.
We collect a list of 532 common object classes (e.g. apple, backpack, book, balloon, avocado; the full list is available in Appendix \ref{supp:full_list}) and use 450 of them for training.
The remaining classes are withheld for testing.
We then construct training prompts by combining two different objects using one of five relationship terms: ``and,'' ``next to,'' ``near,'' ``on side of'' and ``beside,'' producing captions that designate a spatial relationship between two objects, e.g. ``an apple next to an avocado.'' In total we create a training set of over 1M prompts.
In order to compute our object composition reward function (Eq.~\ref{eq:reward-compositionality}), we use UniDet~\cite{zhou2022simple}, an object detector trained on multiple large-scale datasets that supports a wide range of object classes.

\def\thickhline{\noalign{\hrule height.8pt}}
      
\begin{table}[!ht]
\small
\centering 
\vspace{-5pt}
  \begin{tabular}{l c c}
    \toprule
    \multirow{2}{*}{\textbf{Model}} &
      \multicolumn{2}{c}{\textbf{Object Detection Score ($\uparrow$)}}  \\
      \cmidrule(lr){2-3} %
    & Unseen Objects & Seen Objects   \\
    \midrule
    Stable v1.5 & 0.072  & 0.056  \\
    Stable v2 & 0.102 & 0.094 \\

    RAFT & 0.094  & 0.092 \\
    Reward-weighted & 0.136  & 0.152  \\
    Ours & \textbf{0.231} & \textbf{0.221} \\
    \bottomrule
  \end{tabular}%
  \vspace{-5pt}
\caption{\textbf{Compositional Evaluation.} Average detection scores of the objects appearing in the prompts; we report the results on 300 randomly sampled prompts consisting of objects seen by the model during training and another 300 for unseen objects.} 
\vspace{-8pt}
\label{table:compositionality_scores} 
\end{table}

      

 





\begin{table*}[t]

\centering 
\small
  \begin{tabular}{l c c c}
    \toprule
    \multirow{2}{*}{\textbf{Model / Fine-tuning Task}} &
      \multicolumn{3}{c}{\textbf{Evaluation Metic}}  \\
      \cmidrule(lr){2-4} %
    & ImageReward ($\uparrow$)& Object Detection Score ($\uparrow$) & Statistical Parity ($\downarrow$) \\
    \midrule
    Stable v2 & 0.273 & 0.098 & 0.567 \\
    
    Ours -- (ImageReward) & \textbf{0.783} & 0.114  & 0.659 \\
    Ours -- (Compositionality) & 0.304 & \textbf{0.226} & 0.575 \\
    Ours -- (Skintone Diversity) & 0.093 & 0.076 & \textbf{0.479}  \\
    Ours -- Joint & 0.701 & 0.182 & 0.499  \\
    \bottomrule
  \end{tabular}%

\vspace{-5pt}
\caption{\textbf{Joint Optimization.} We experiment with jointly optimizing a single model to satisfy three separate reward functions. Comparing with the original baseline model, we see that our jointly optimized model is able to satisfy all three objectives, achieving over 80\% (relative) performance of the individually fine-tuned models across all three evaluation metrics simultaneously.} 
\label{table:joint_training} 
\vspace{-15pt}
\end{table*}

We present qualitative and quantitative results in Figure~\ref{fig:sd2_ours_compositionality} and Table~\ref{table:compositionality_scores}.
To evaluate generalizability, we also generate samples with our fine-tuned model on 300 randomly sampled prompts from both unseen and seen objects. Our trained model adheres better to compositional constraints in text captions when compared to SDv2, and the learned compositional abilities also generalize to unseen objects.

\subsection{Multi-reward Joint Optimization}
\label{sec:4.4multireward}

As detailed in Algorithm~\ref{alg:multi-reward}, we also perform multi-reward RL with all three reward functions jointly, aiming to improve the model performance on all three tasks simultaneously. We compare the jointly-trained model with the base model and the models fine-tuned for each individual task. The quantitative results are shown in Table~\ref{table:joint_training}, with more qualitative results available in Appendix \ref{supp:additional_qualit}. Following the same evaluation setting, we test the models on multiple datasets for each metric
and report the average scores. 

\looseness=-1
While the best score for each metric is achieved by the model fine-tuned specifically for that task, our jointly-trained model is able to satisfy over 80\% (relative) performance of the individually fine-tuned models across all three metrics simultaneously. In addition, it significantly outperforms the original base model on all tasks. 

\paragraph{Alignment Tax} We observed degraded performance for individually fine-tuned models on some of the tasks that the models were not fine-tuned for. For example, the model optimized for human preference exhibits a significant regression on statistical parity, indicating a drastic drop in skintone diversity. Similarly, the model optimized for skintone diversity degrades in terms of human preference as compared to the base model.  This is akin to the ``alignment tax" issue that has been observed during RLHF
fine-tuning procedure of LLMs~\cite{ouyang2022training, askell2021general}.
Specifically, when models are trained with a reward function that is only concerned with one aspect of images, it may learn to neglect sample quality or overall diversity of outputs.
Our jointly fine-tuned model, in contrast, is able to mitigate the alignment tax issue by incorporating multiple diverse reward functions during fine-tuning, thereby maintaining performance on all tasks in question.
\vspace{-7pt}
\section{Conclusion}
\vspace{-3pt}

\looseness=-1
We present a scalable RL training framework for directly optimizing diffusion models with arbitrary downstream objectives, including distribution-based reward functions. We conducted large-scale multi-task fine-tuning to improve the general performance of an SDv2 model in terms of human preferences, fairness, and object composition simultaneously, and found that joint training also mitigated the alignment tax issue common in RLHF. By evaluating our trained model against several baseline models on diverse out-of-domain test sets, we demonstrated our method's generality and robustness. We hope our work inspires future research on targeted tuning of diffusion models, with potential future topics including addressing more complex compositional relationships and mitigating bias along other social dimensions.



\newpage

{
    \small
    \bibliographystyle{ieeenat_fullname}
    \bibliography{main}
}

\newpage
\appendix
\onecolumn

{\large\textbf{Appendix}}
\\ \\
This appendix is structured as follows:

\begin{itemize}
\setlist[itemize]{leftmargin=0.5em}
\setlength\itemsep{0.3em}
    \item In Appendix \ref{supp:hyperparameters}, we provide more details of our experimental setup, including hyperparameters and baselines.
    \item In Appendix \ref{supp:additional_qualit}, we provide additional qualitative results and comparison of our method with the baselines.

\item In Appendix \ref{supp:human_eval_temp}, we provide evaluation guidelines and templates used for collecting human rating.
\item In Appendix \ref{supp:additional_human_eval}, we provide additional human evaluation results for skintone diversity and compositionality.
\item In Appendix \ref{supp:training_curve}, we provide the training curves of all online-learning methods (including ours and other baselines) to \\ demonstrate the training progress and convergence time. 
\item In Appendix \ref{supp:reward_hacking}, we illustrate the issue of reward hacking and provide visual examples.
\item In Appendix \ref{supp:pretraining_effect}, we provide an ablation study on the effect of pretraining dataset.
\item In Appendix \ref{supp:full_list}, we provide complete lists of 100 occupations for skintone diversity evaluation and 532 objects for training and evaluating the compositionality skill of the models.

\end{itemize}

\section{Experiment Details and Hyperparameters}
\label{supp:hyperparameters}
All our experiments including baseline methods training were conducted on 128 80GB A100 GPUs. If a pretraining dataset is required, all fine-tuning methods use the same 12M subset of LAION-5B~\cite{schuhmann2022laion5b} filtered by the aesthetic
score predictor with a threshold of 6.  For optimization, we use the AdamW optimizer~\cite{loshchilov2019decoupled} with $\beta_1 = 0.9, \beta_2 =
0.999, \epsilon=1e-8$ and a weight decay of $1e-2$ for all the experiments. For inference, we run the diffusion process with
50 steps for each image with DDIM~\cite{song2022denoising} noise scheduler.
We use the default guidance scale of 7.0 for classifier-free
guidance~\cite{ho2022classifierfree}.

\paragraph{Implementation Details} For our RL fine-tuning experiments, we collect 16x128 samples per training iteration, with 50 samplings steps using DDIM scheduler~\cite{song2022denoising}. We randomly sample 5 training timesteps and perform a gradient update across all the samples in the batch for each of the timesteps, resulting in 5 gradient updates per iteration. We use a small clip range of $1e-4$ for all the experiments.

\paragraph{Baseline Details} For the baseline methods including ReFL~\cite{xu2023imagereward}, Reward-weighted~\cite{lee2023aligning}, RAFT~\cite{dong2023raft} and DRaFT~\cite{clark2023directly}, we refer to the original implementation for the suggested hyperparameters and report our experiment details in Table \ref{table:hyperparameters}. We use the same training set for all the baseline models training and fine-tune them until convergence. Since the experiments involve million-sized training prompts, for reward-weighted approach, instead of pre-generating the samples and storing the dataset offline, we generate the samples on the fly during training using the original SDv2 model and re-weigh them according to the reward values for fine-tuning. Following Xu~\etal~\cite{xu2023imagereward}, we also map the reward values to the range of $[0, 1]$ using min-max normalization.

We note that DRaFT imposes a high memory burden by directly back-propagating the gradient from the reward model through the sampling process of diffusion model, allowing for a much smaller batch size compared to other optimization methods. We implement DRaFT-LV, which claimed to be the most efficient DRaFT variant. 
\begin{table}[H]
\small
\centering 
\vspace{-4pt}
\resizebox{\textwidth}{!}{
  \begin{tabular}{l| c |c| c| c |c}
    \toprule
\textbf{Hyperparameter} & \textbf{ReFL} & \textbf{Reward-weighted} & \textbf{RAFT} & \textbf{DRaFT}  & \textbf{Ours} \\
    \hline
    Learning Rate & 1e-5 & 1e-5 & 3e-6 & 5e-5 & 2e-6  \\
    Batch Size (Per GPU) & 12 & 16 & 32 & 3 & 16  \\
    Pretraining Batch Size (Per GPU) &  12 & 16 & 32 & - & 16  \\
    Sampling Scheduler & DDIM & DDIM & DDIM & DDIM & DDIM \\
    Sampling Steps & 40 & 50 & 50 & 50 & 50 \\
     \hline
    Method Specific 
    & \makecell{ $\phi=ReLU$ \\ $[T1,T2]=[1,10]$ \\
    $\lambda=1e-3$}
    & $\beta=0.5$
    & Acceptance ratio: 1/24
    & \makecell{ LoRA rank: 32 \\ t\textsubscript{truncate}=1}
    & \begin{tabular}{@{}c@{}}clip range: 1e-4 \\ Training timesteps: 5 \end{tabular}  \\

    \bottomrule
    
  \end{tabular}%
  }
\caption{\textbf{Training Hyperparameters.} We report the  hyperparameters used in different experiments, where \textit{method-specific} indicates the hyperparameters specific to each individual method.} 
\vspace{-8pt}
\label{table:hyperparameters} 
\end{table}

\section{Additional Qualitative Results}
\label{supp:additional_qualit}
We provide additional qualitative results in this section, including results from the models that were trained with single rewards (i.e., ImageReward~\cite{xu2023imagereward}, compositionality reward and skintone diversity reward), as well as the results from our model that was jointly trained with all three rewards simultaneously. 

\subsection{Results on Human Preference Fine-tuning}
\begin{figure}[H]
  \centering
  \vspace{-8pt}
   \includegraphics[width=\linewidth]{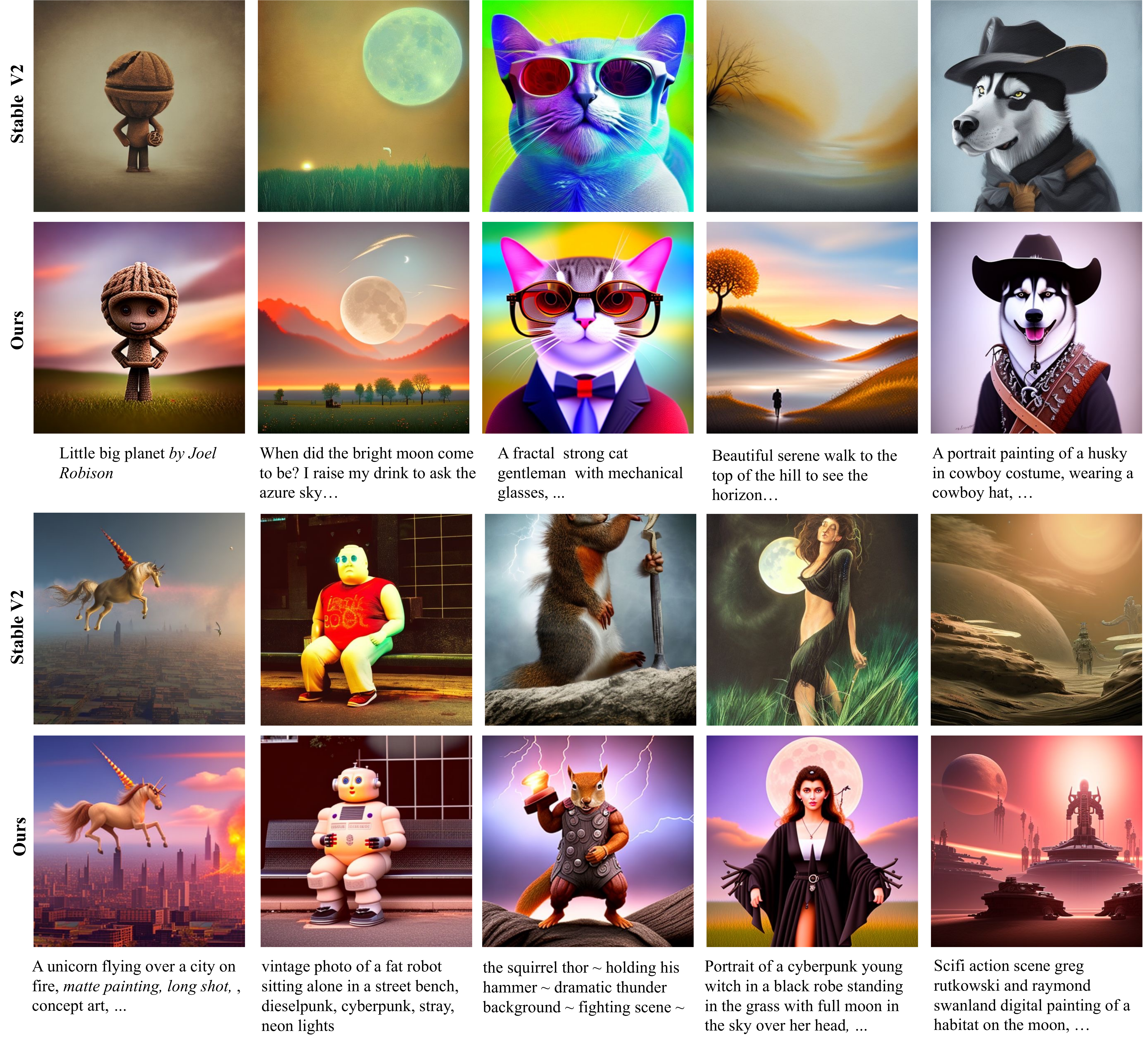}
   \vspace{-9pt}
   \caption{\textbf{Qualitative Comparison of SDv2 and Our Fine-tuned Model}. All images are generated using real-user prompts from DiffusionDB~\cite{wangDiffusionDBLargescalePrompt2022} dataset with the same random seeds. Our outputs are better aligned with human aesthetic preferences, favoring  finer details, focused composition, vivid colors, and high contrast.}
   \label{fig:sd2_ours_2x4}
\end{figure}

\begin{figure}[H]
  \centering
  \vspace{-15pt}
   \includegraphics[width=\linewidth]{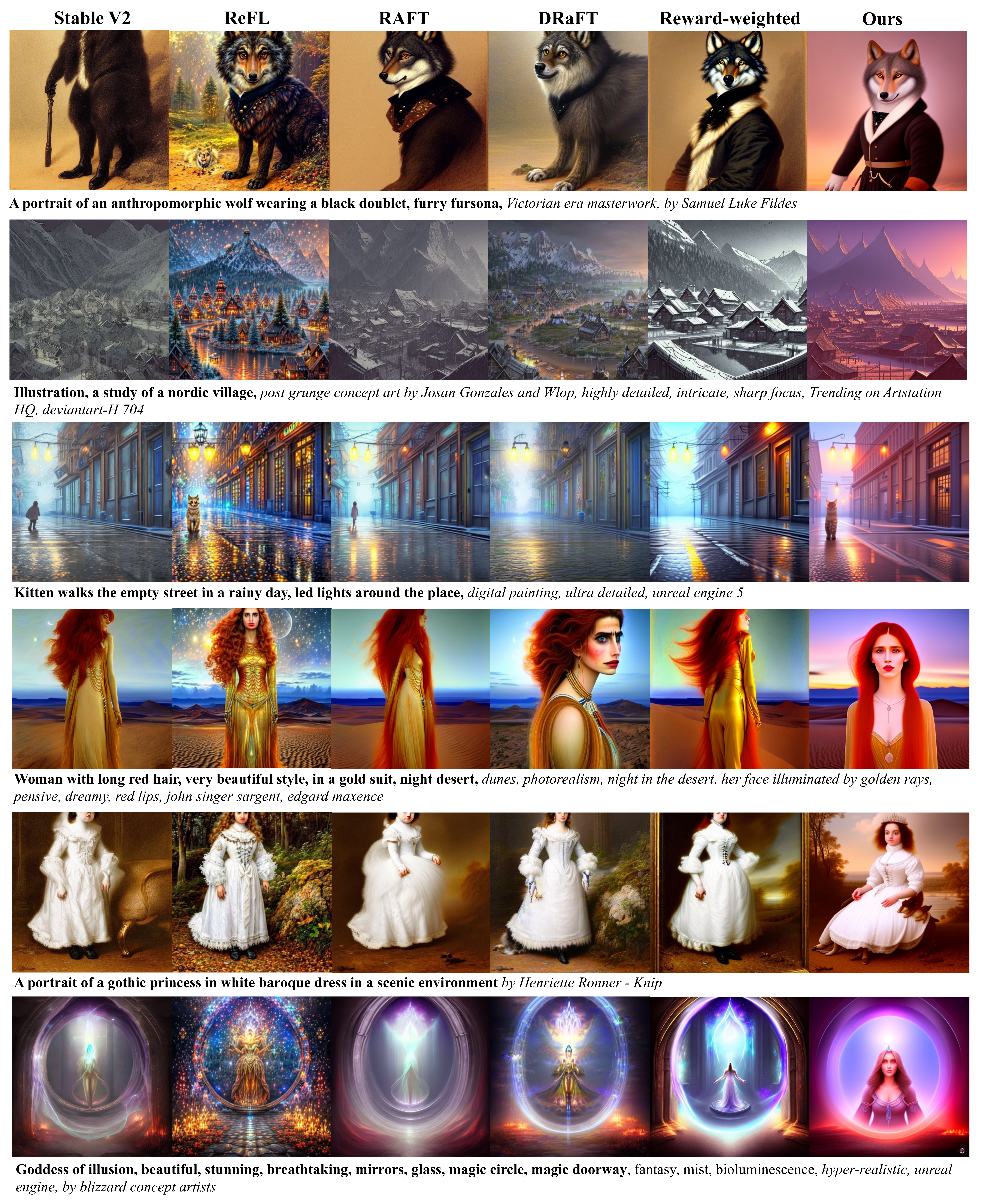}
   \vspace{-15pt}
   \caption{\textbf{Additional Qualitative Comparison Results}. We compare our fine-tuned model with other reward fine-tuning methods on real-user prompts from DiffusionDB~\cite{wangDiffusionDBLargescalePrompt2022} dataset. All images are generated using the same random seeds. }
   \label{fig:im_reward_vis_addtional1}
\end{figure}

\begin{figure}[H]
  \centering
  \vspace{-15pt}
   \includegraphics[width=\linewidth]{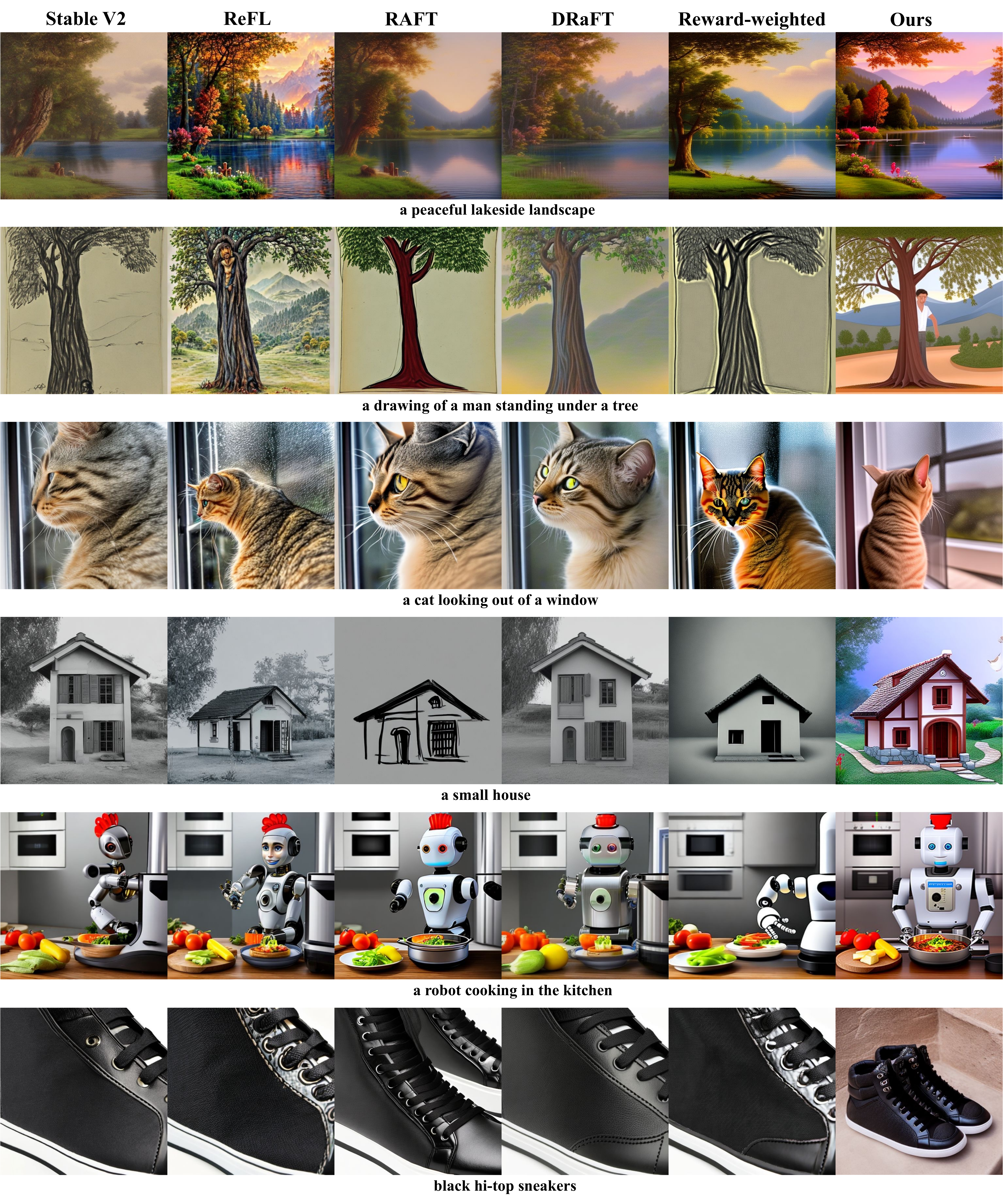}
   \vspace{-16pt}
   \caption{\textbf{Additional Qualitative Comparison on Out-of-domain Test Sets}. We compare our fine-tuned model with other reward fine-tuning methods on PartiPrompts~\cite{yu2022scaling} dataset. Our model generates samples with higher aesthetic quality and better image-text alignment compared to other baseline models. All images are generated using the same random seeds.   }
   \label{fig:im_reward_vis_addtional2}
\end{figure}

We show the visual samples from our model fine-tuned with ImageReward~\cite{xu2023imagereward} on real-user prompts in Figure \ref{fig:sd2_ours_2x4}. We also provide more qualitative comparison of our model with other reward optimization methods in Figure \ref{fig:im_reward_vis_addtional1}. More results on the out-of-domain test set PartiPrompts~\cite{yu2022scaling} are availble in Figure \ref{fig:im_reward_vis_addtional2}. Our trained model generates more visually appealing images compared to the base SDv2 model, and it generalizes well to out-of-domain test sets with unseen text prompts that have a different distribution from that of the training prompts.

\subsection{Results on Optimizing Diversity}
\begin{figure}[H]
  \centering
  \vspace{-6pt}
   \includegraphics[width=.8\linewidth]{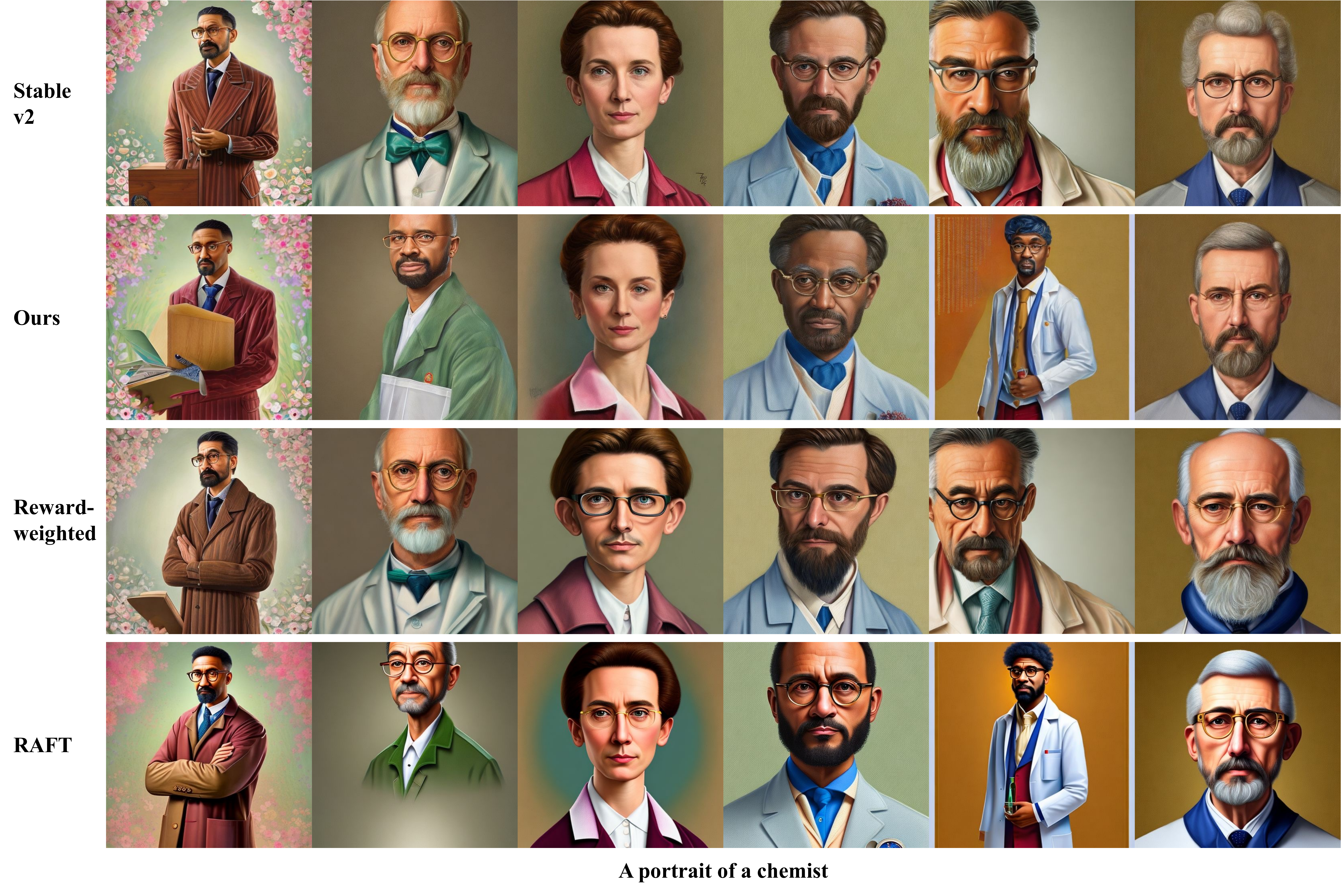}
   \includegraphics[width=.8\linewidth]{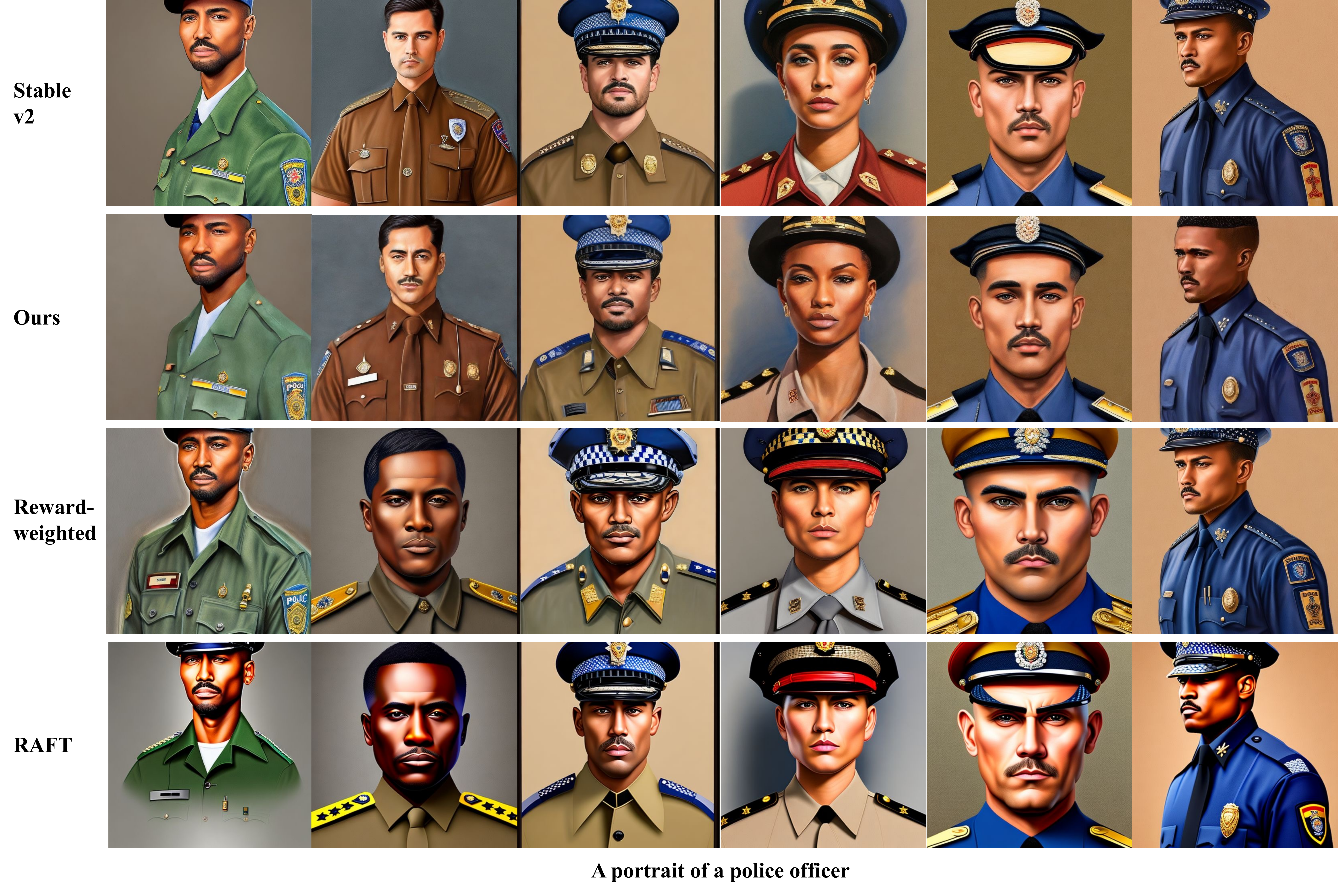}
   \vspace{-6pt}
   \caption{\textbf{Skintone Diversity Visualization}. We compare our model that was fine-tuned with skintone diversity  reward with other baseline models. All images are generated using the same random seeds. We note that while RAFT also improves the skintone diversity of the output samples, it is prone to overfitting and generates over-saturated samples with decreased realism (e.g. the portraits of police officers in the second example).}
   \label{fig:vis_addtional_skintone}
\end{figure}

We provide more qualitative results of our model fine-tuned with skintone diversity reward in Figure \ref{fig:vis_addtional_skintone}. Our trained model effectively mitigates the inherent bias and stereotypes in the base SDv2 model with increased skintone diversity in the generated human samples. 

\subsection{Results on Optimizing Compositionality}
\begin{figure}[H]
  \centering
   \includegraphics[width=\linewidth]{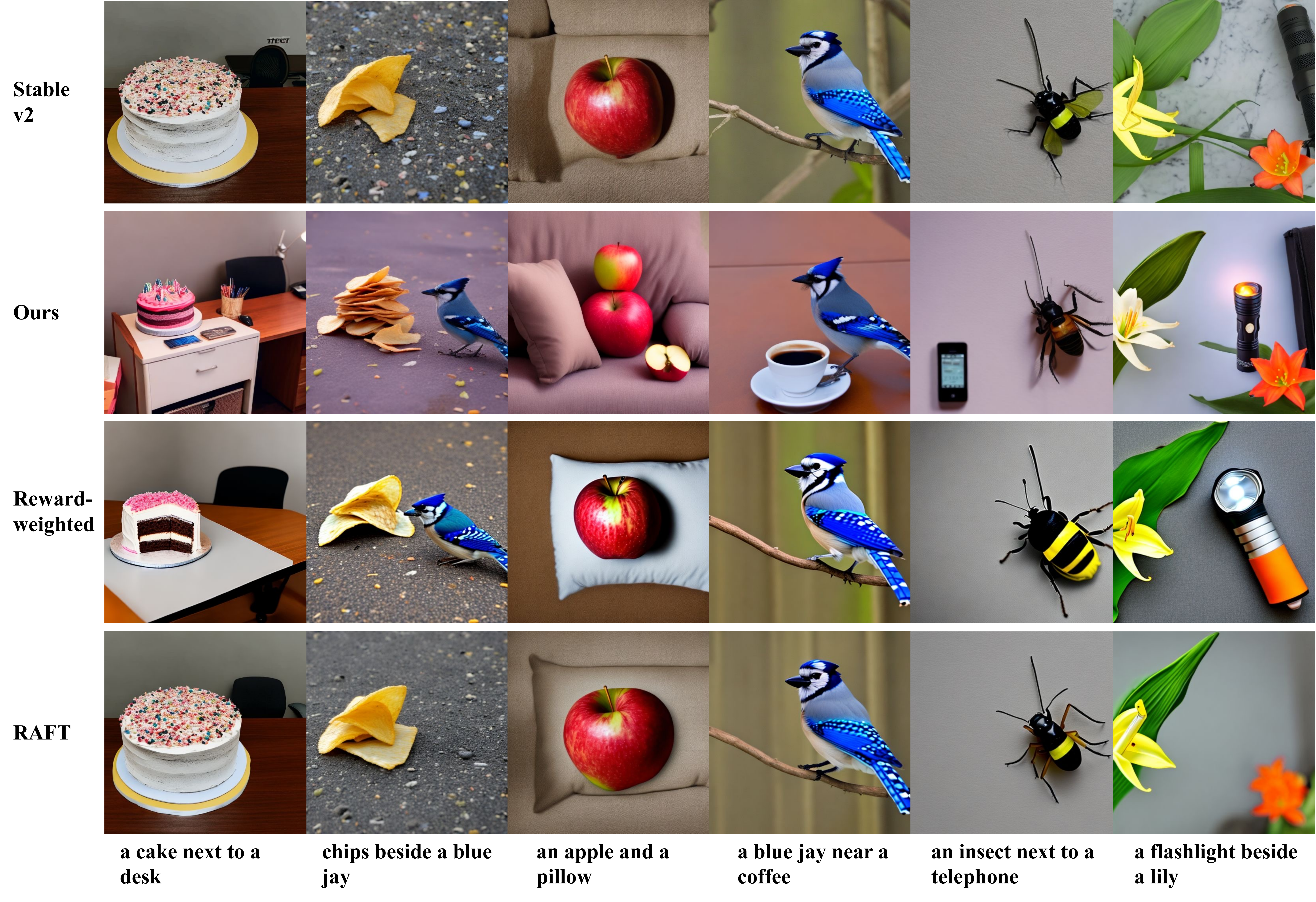}
   \caption{\textbf{Object Composition Visualization}. We compare our model that was fine-tuned with object composition reward with other baseline models. All images are generated using the same random seeds. }
   \label{fig:vis_addtional_composition}
\end{figure}

We provide more qualitative results of our model fine-tuned with object composition reward in Figure \ref{fig:vis_addtional_composition}. Our fine-tuned model demonstrates improved compositional skills compared to the base SDv2 model and other baseline models.

\subsection{Results on Multi-reward Joint Optimization}
\begin{figure}[H]
  \centering
  \vspace{-15pt}
   \includegraphics[width=\linewidth]{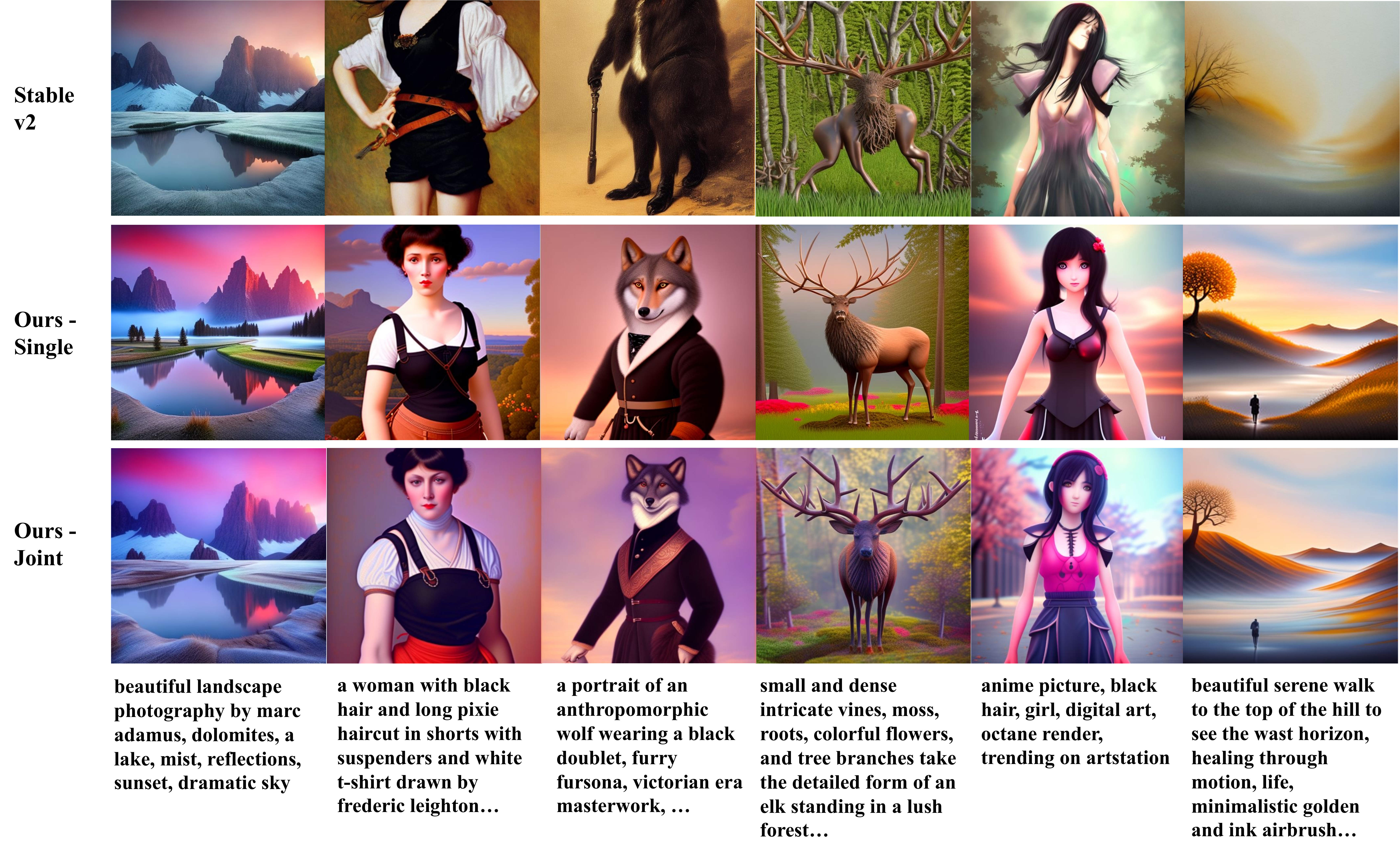}
   \vspace{-10pt}
   \caption{\textbf{Visualization of Jointly-optimized Model on Real-user Prompts}. We show the results from our jointly-fined-tuned model on real-user prompts from DiffusionDB~\cite{wangDiffusionDBLargescalePrompt2022} dataset. Our jointly-trained model generates more aesthetically pleasing images compared to the base SDv2 model.}
   \label{fig:joint_im}
\end{figure}

\begin{figure}[H]
  \centering
  \vspace{-15pt}
   \includegraphics[width=\linewidth]{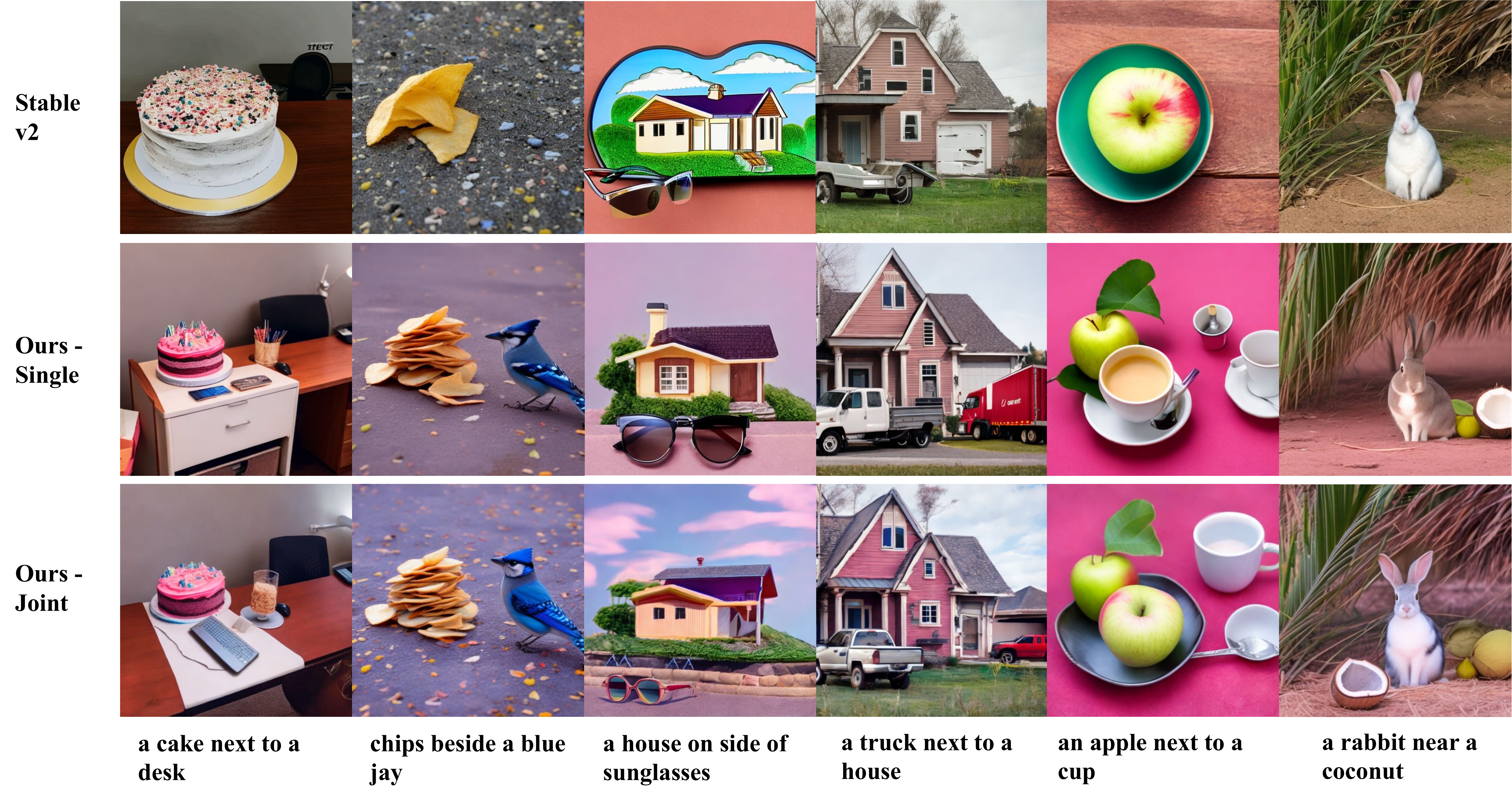}
   \vspace{-10pt}
   \caption{\textbf{Visualization of Jointly-optimized Model on Object Composition Prompts.} Our jointly-trained model generates samples with improved compositionality compared to the base SDv2 model.}
   \label{fig:joint_composition}
\end{figure}

\begin{figure}[H]
  \centering
  \vspace{-15pt}
   \includegraphics[width=\linewidth]{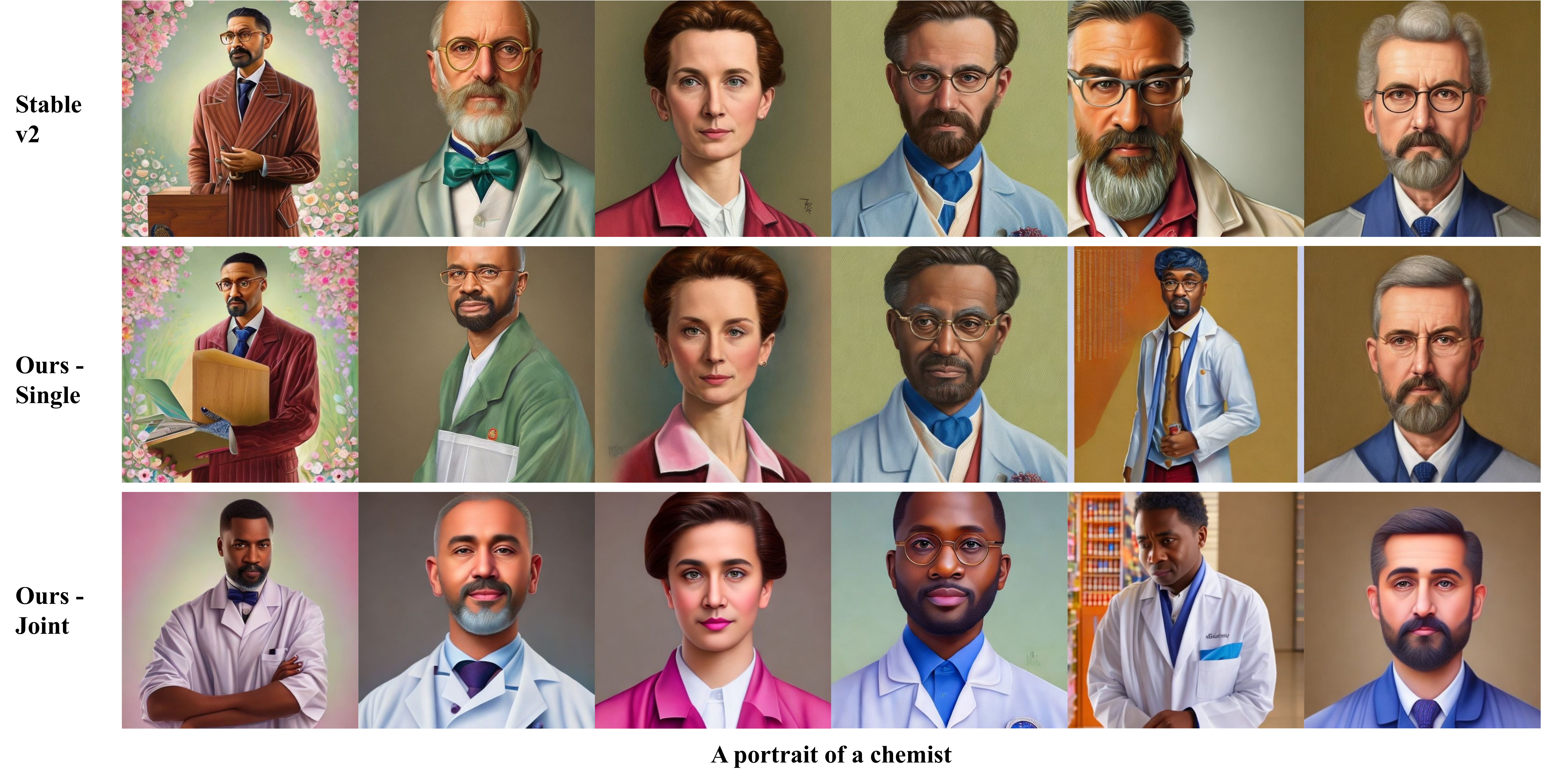}
   \includegraphics[width=\linewidth]{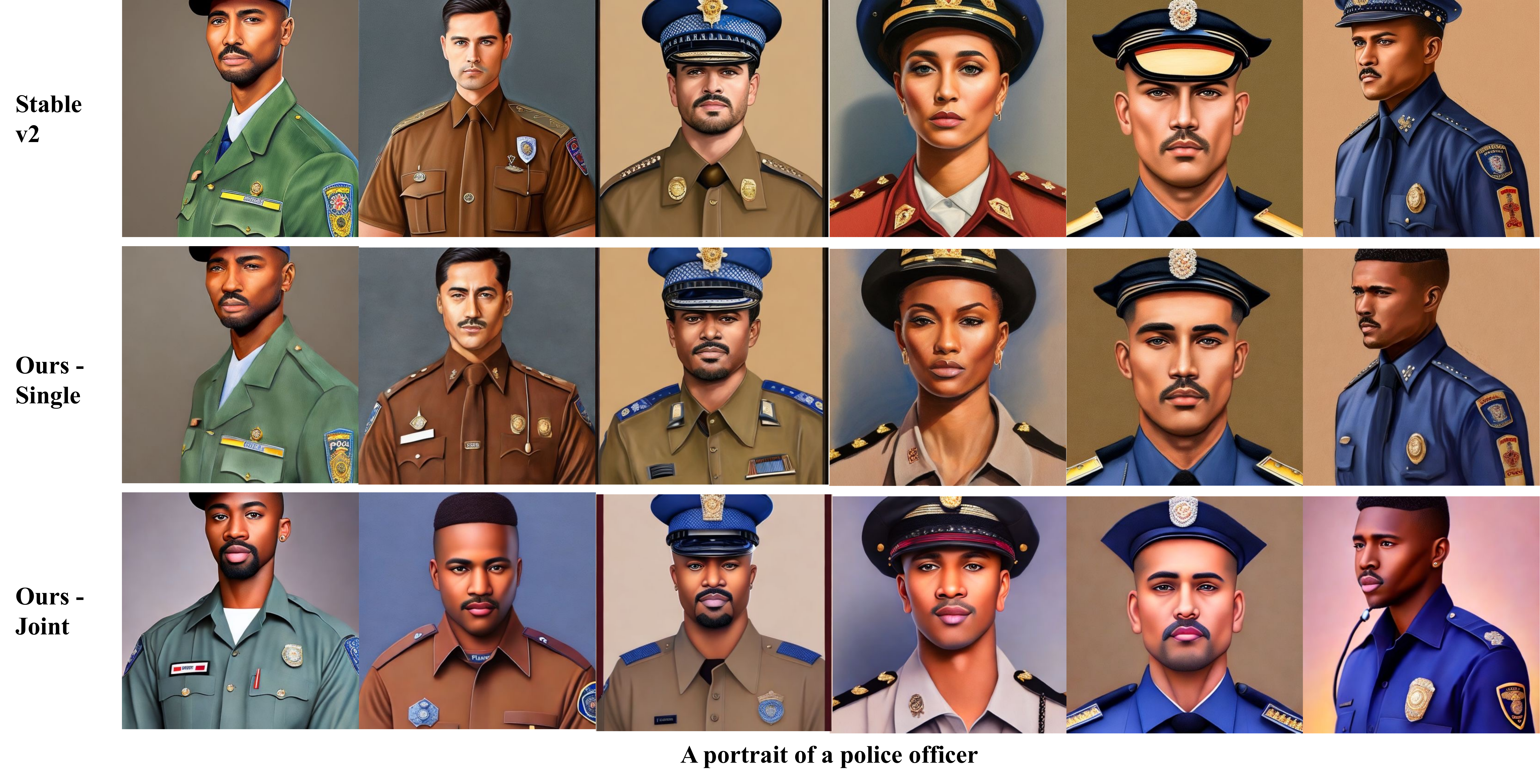}
   \vspace{-6pt}
   \caption{\textbf{Visualization of Jointly-optimized Model on Occupation Prompts}. We show the results from our jointly-fined-tuned model on occupations prompts for skintone diversity evaluation. Our jointly-trained model has greatly reduced the inherent bias in the base SDv2 model and generates human samples with more diverse skintone. Compared to single-reward training, we also observe the additional increase in the aesthetic quality of the samples from our jointly-trained model.}
   \label{fig:joint_skintone}
\end{figure}

Next, we show more qualitative results from our jointly-fine-tuned model (with all three rewards simultaneously) on multiple test sets: DiffusionDB~\cite{wangDiffusionDBLargescalePrompt2022} (Figure \ref{fig:joint_im}), object composition (Figure \ref{fig:joint_composition}) and occupation prompts (Figure \ref{fig:joint_skintone}). We demonstrate that our jointly-trained model has quite significant improvement over the base SDv2 model in terms of all three objectives: human preferences, skintone diversity and object composition. We further note that since joint training utilizes multiple reward signals (including ImageReward which reflects human preferences) during training, for portraits of occupations, we also observe additional increase in the aesthetic quality of the samples compared to single-reward training which optimizes for the skintone diversity only; see Figure \ref{fig:joint_skintone}.

\section{Human Evaluation Templates}
\label{supp:human_eval_temp}
We provide the detailed human evaluation guidelines document that were used to train our hired human labelers in section   \ref{subsec:eval_guidelines}, including the judging criteria and concrete examples for making trade-offs in order to help the evaluators better understand the task and make fair judgments. We use the annotation documents from ImageReward~\cite{xu2023imagereward} as a reference.
We also show our evaluation UI interface in section \ref{subsec:eval_interface}.

\subsection{Evaluation Criteria and Guidelines }
\label{subsec:eval_guidelines}
You will be given a number of prompts/queries and there are several AI-generated images according to the prompt/query. Your annotation requirement is to evaluate these images in terms of \textbf{Image Fidelity}, \textbf{Relevance to the Query}, and \textbf{Aesthetic Quality}. Below are more details on each of the three mentioned factors.

\subsubsection{Image Fidelity}

\begin{figure}[H] \centering
  \includegraphics[width=1.6in,height=1.6in]{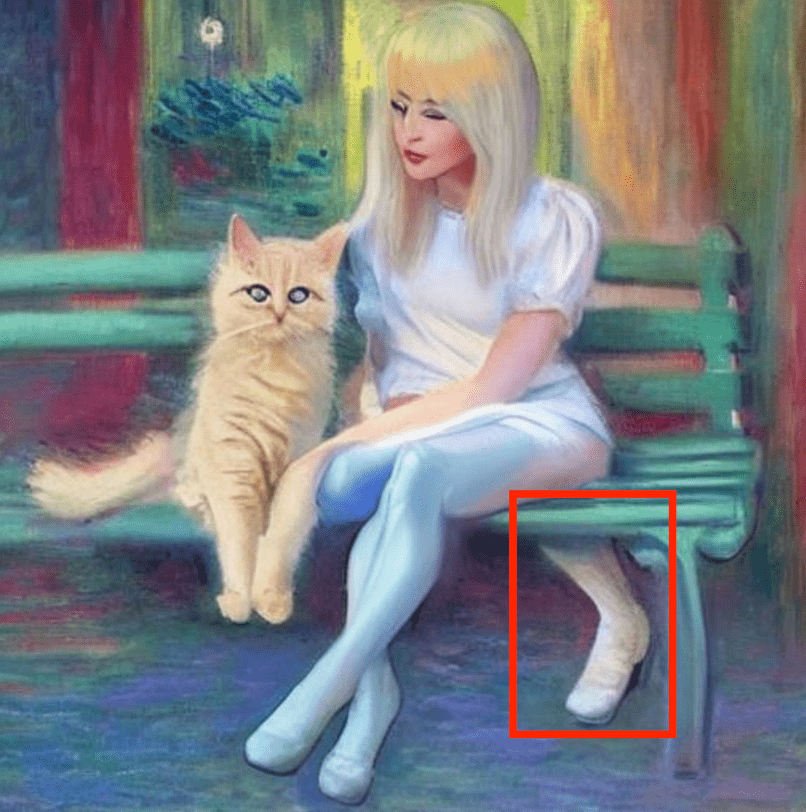}\hspace{1em}%
  \includegraphics[width=1.6in,height=1.6in]{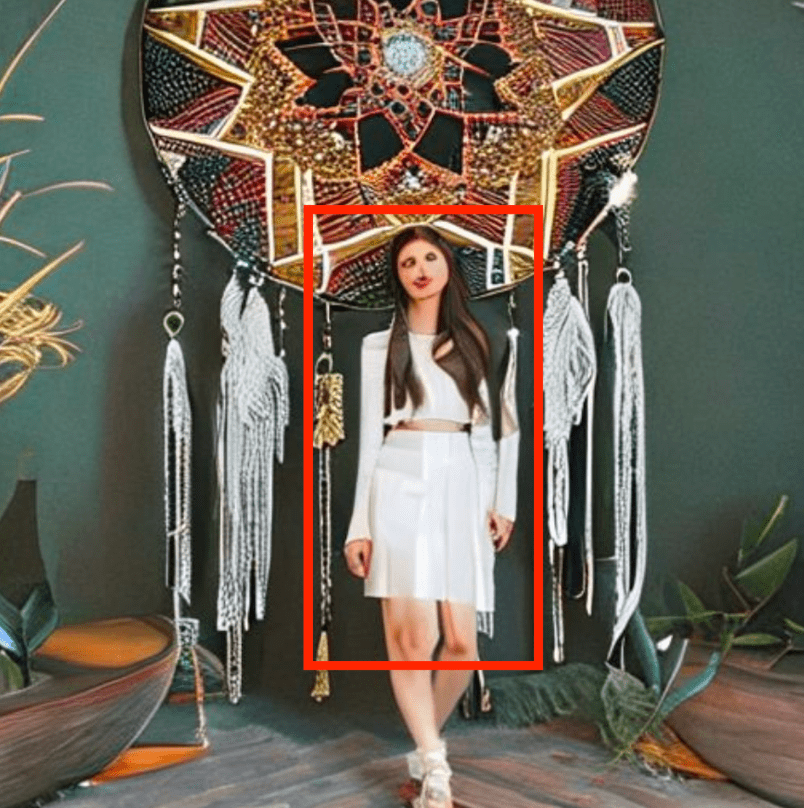}\hspace{1em}%
  \includegraphics[width=1.6in,height=1.6in]{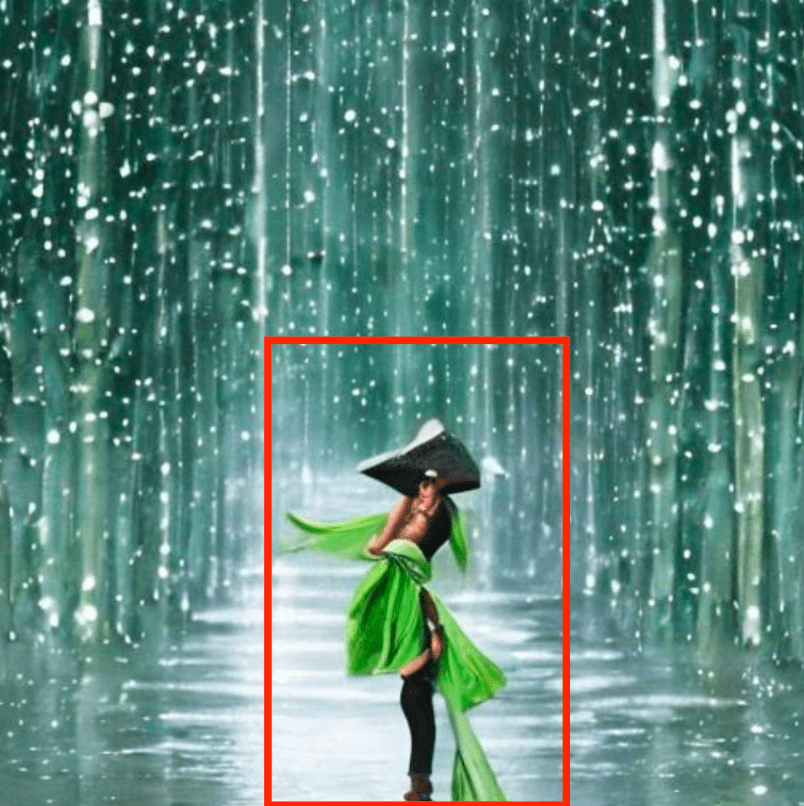}\hspace{1em}%
  \includegraphics[width=1.6in,height=1.6in]{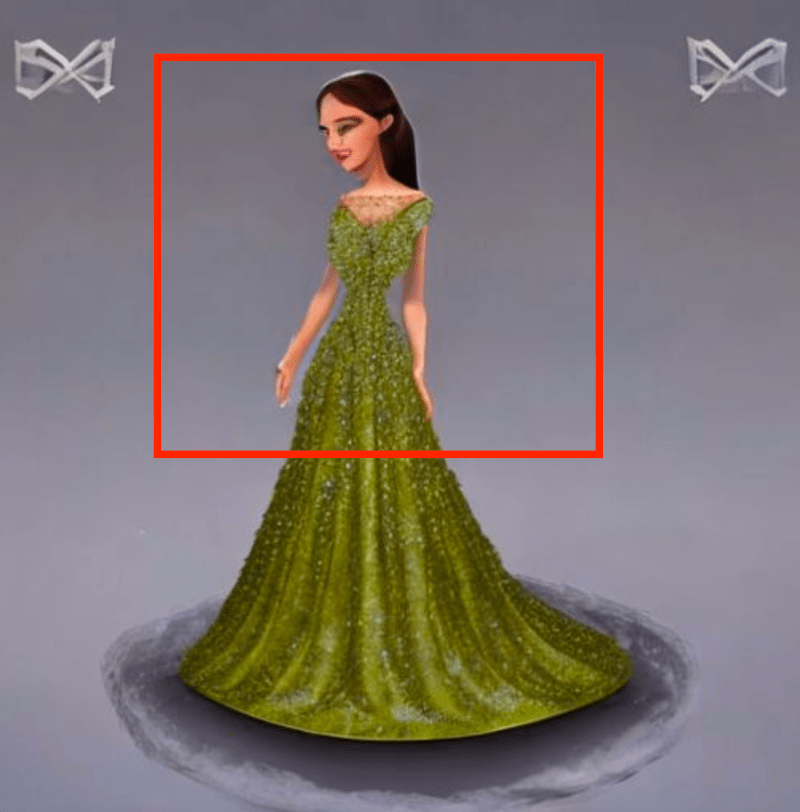}\hspace{1em}%
  \includegraphics[width=1.6in,height=1.6in]{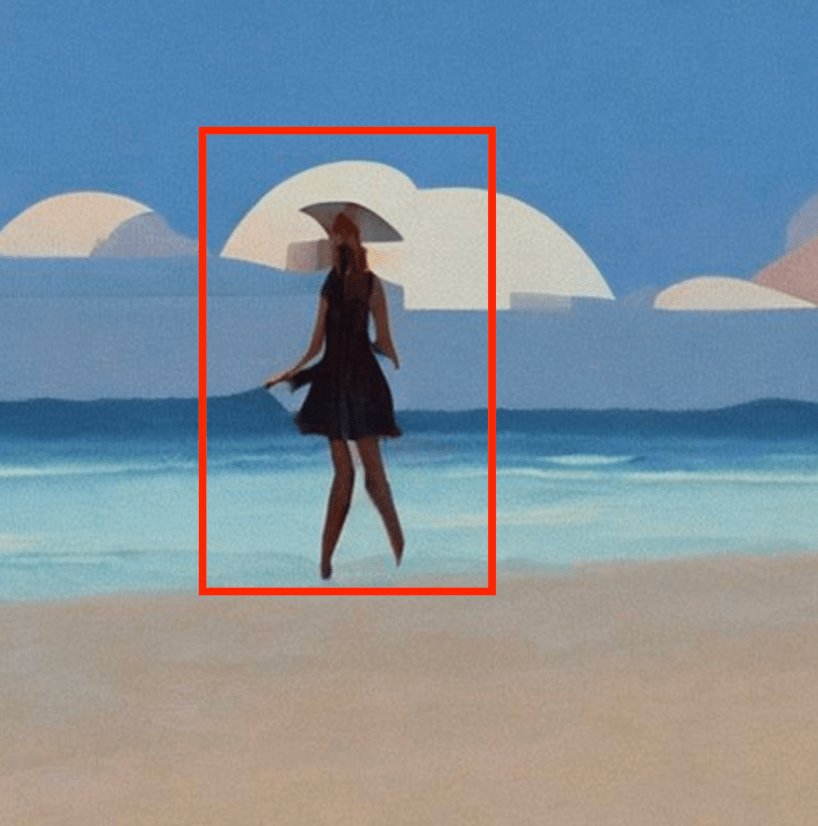}\hspace{1em}%
  \includegraphics[width=1.6in,height=1.6in]{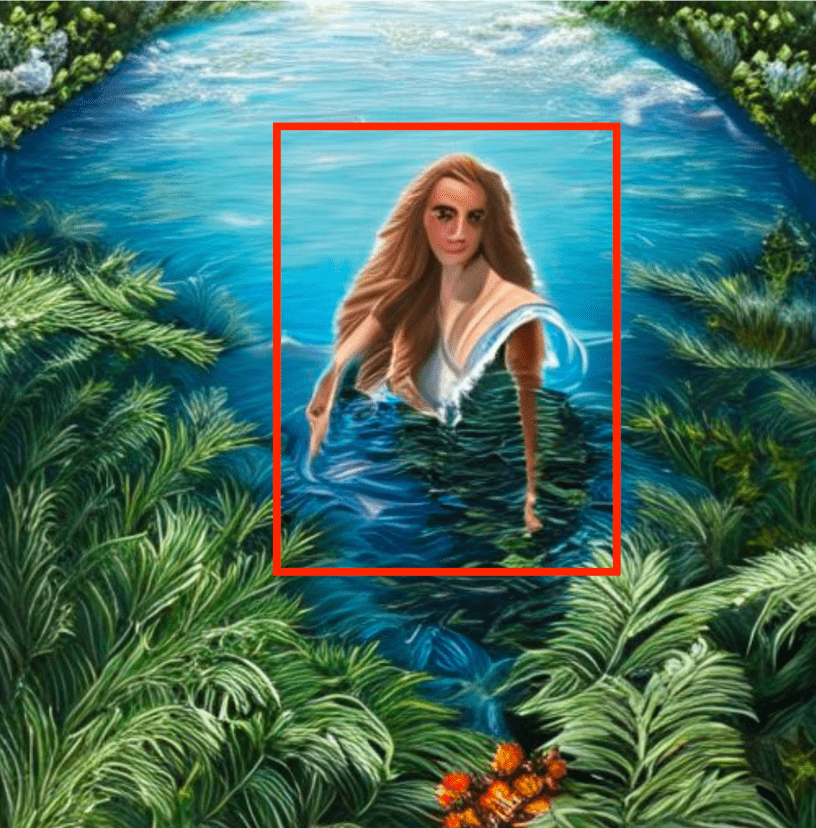}\hspace{1em}%
   \includegraphics[width=1.6in,height=1.6in]{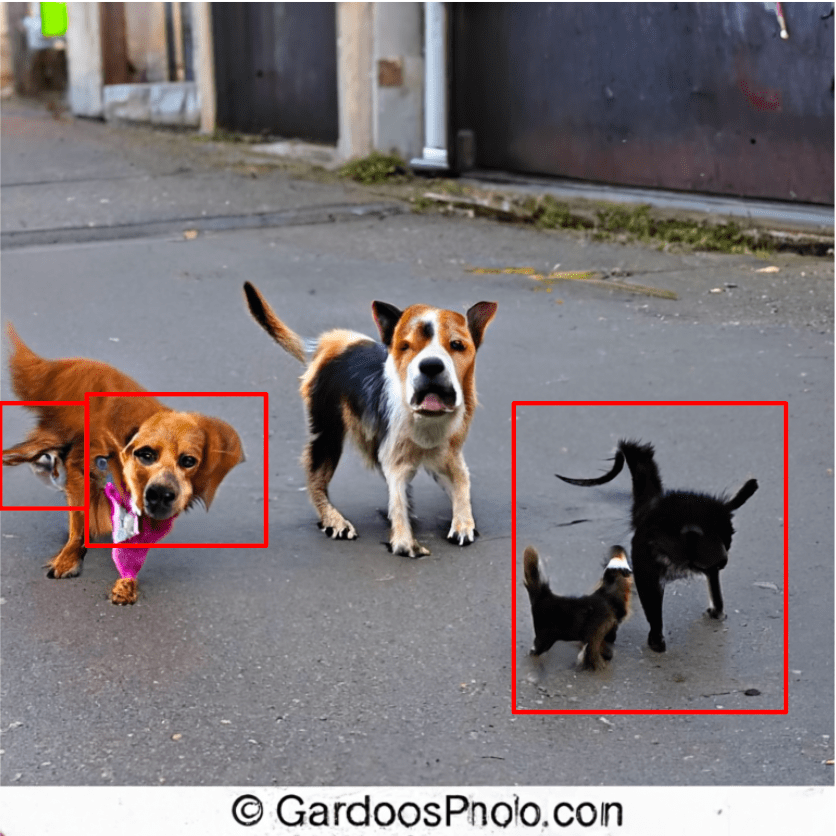}\hspace{1em}%
    \includegraphics[width=1.6in,height=1.6in]{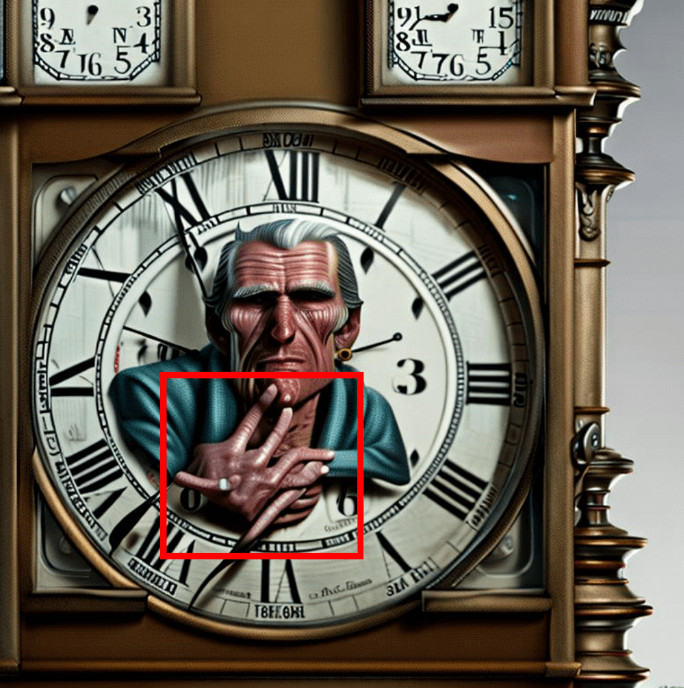}\hspace{1em}%
    \caption{\textbf{Examples of Low-fidelity Generation.} Note that these generated images have incorrect details with faces or body parts of human and animals, and would likely cause psychological discomfort. They should be ranked with lower-preference.
}
  \label{fig:low_fidelity_examples}
    
\end{figure}

\textbf{Definition: }The generated image should be true to the shape and characteristics of the object, and not generated haphazardly. 
Some examples of low-fidelity images are: 
\begin{itemize}
    \item Dogs should have four legs and two eyes, generating an image with extra / fewer legs or eyes is considered low-fidelity.
    \item ``Spider-Man"" (or human) should only have two arms and five fingers each. Generating extra arms / fingers is considered low-fidelity. 
    \item ``Unicorn" should only have one horn, generating an image with multiple horns is considered low-fidelity. 
    \item People eat noodles with utensils instead of grabbing them with their hands, generating an image of someone eating noodles with their hands is considered low-fidelity.
\end{itemize}
See Figure \ref{fig:low_fidelity_examples} for examples of low-fidelity generation.
Images of low fidelity should be ranked as low preference. 

\subsubsection{Relevance to the Query}
\textbf{Definition}: the generated image should match the text in the query. Another term used for``Relevance" is ``Text-alignment".
Some examples of inconsistent image generation are: 
\begin{itemize}
\item The subject described in the text does not appear in the image generated, for example, ``A cat dressed as Napoleon Bonaparte" generates an image without the word ``cat". 
\item The object properties generated in the image are different from the text description, for example, generating an image of “a little girl sitting in front of a sewing machine” with a boy (or many little girls) is incorrect.
\end{itemize}

\begin{figure}[H] \centering
   \subcaptionbox{``A cucumber beside a peach
"\label{low_rel1}}{\includegraphics[width=1.6in,height=1.6in]{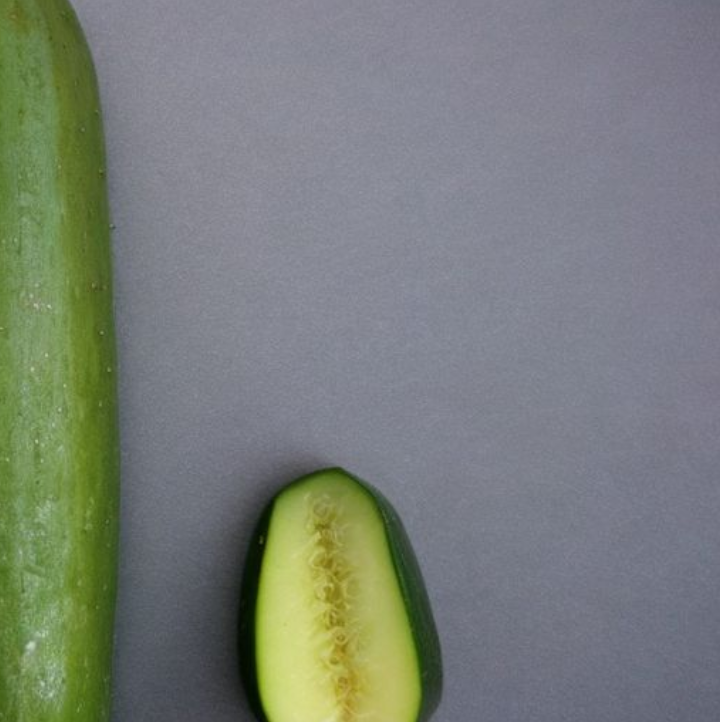}}\hspace{1em}%
  \subcaptionbox{``A box and a canoe"
\label{low_rel2}}{\includegraphics[width=1.6in,height=1.6in]{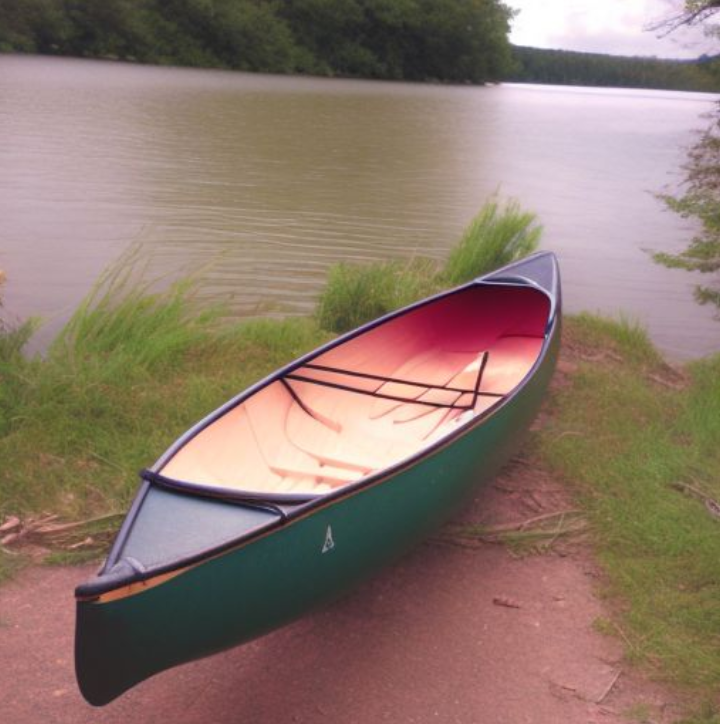}}\hspace{1em}%
  \subcaptionbox{``set of 2 canvas paintings"\label{low_rel3}}{\includegraphics[width=1.6in,height=1.6in]{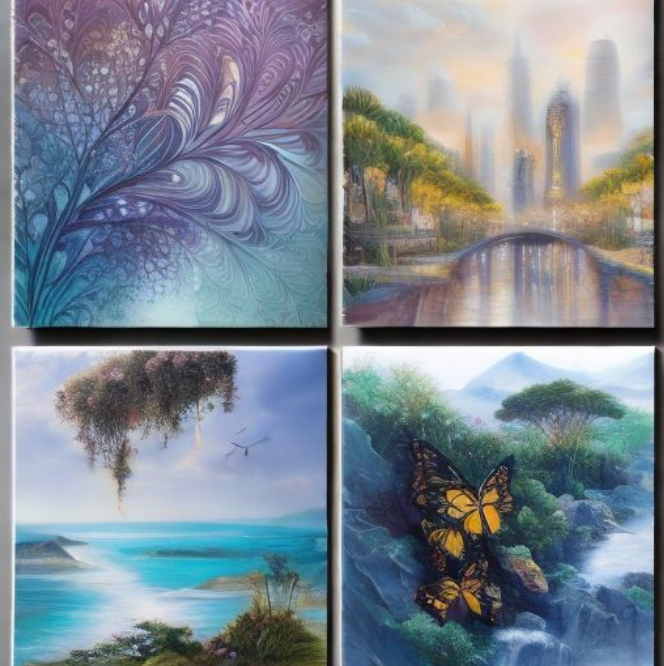}}\hspace{1em}%
  \subcaptionbox{``sip and paint at home date night"\label{low_rel4}}{\includegraphics[width=1.6in,height=1.6in]{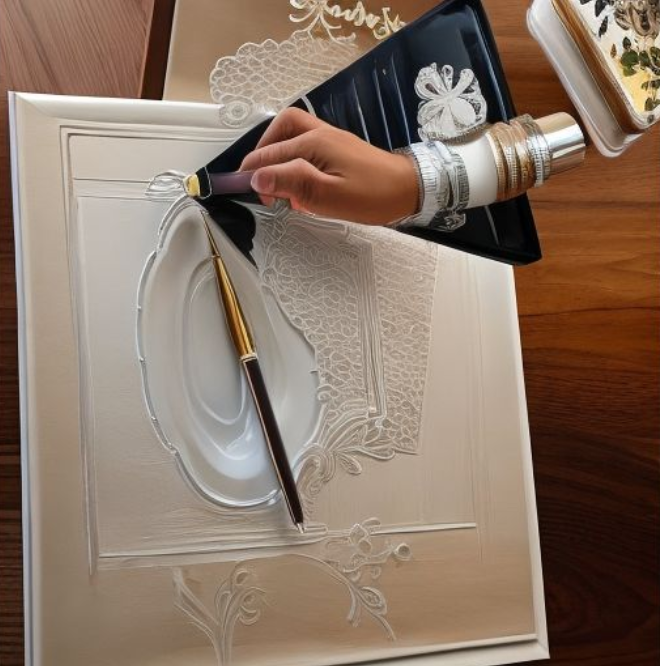}}\hspace{1em}%
  \subcaptionbox{``Matching wallpaper for two best friends"\label{low_rel5}}{\includegraphics[width=1.6in,height=1.6in]{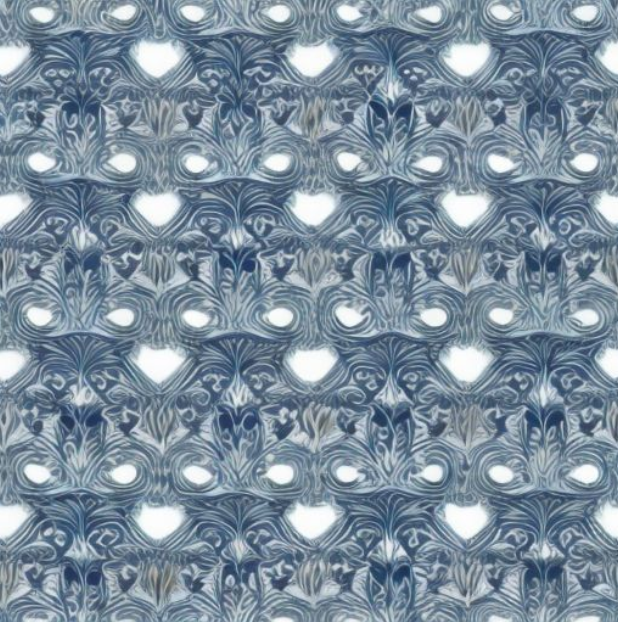}}\hspace{1em}%
  \subcaptionbox{``Underwater congress art"\label{low_rel6}}{\includegraphics[width=1.6in,height=1.6in]{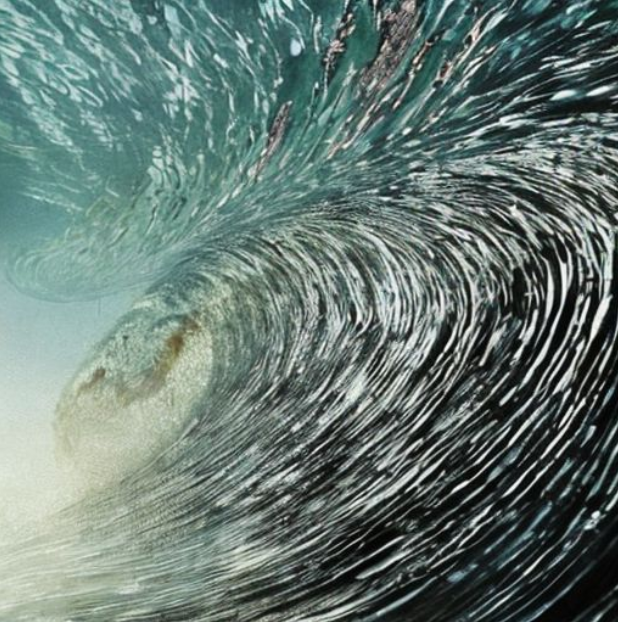}}\hspace{1em}%
  \subcaptionbox{``Shoe design sketches draw
"\label{low_rel7}}{\includegraphics[width=1.6in,height=1.6in]{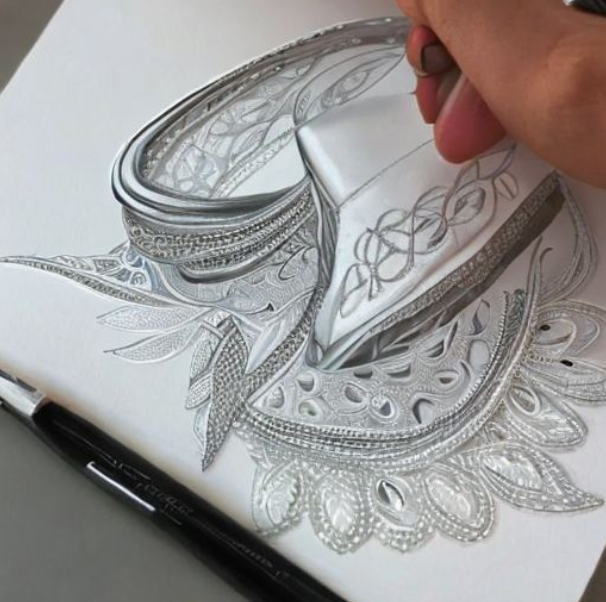}}\hspace{1em}%
  \subcaptionbox{``Black cat minimalist art
"\label{low_rel8}}{\includegraphics[width=1.6in,height=1.6in]{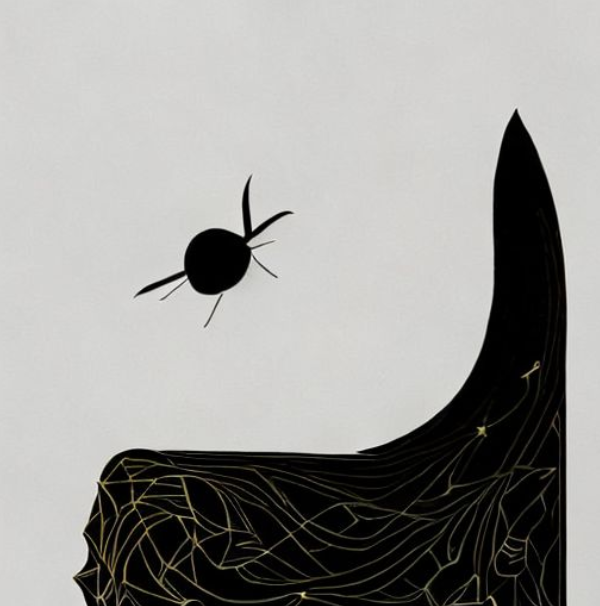}}\hspace{1em}%

    \caption{\textbf{Examples of Generation with Low-relevance to the Text Prompts.} They should be ranked with lower-preference.
}
  \label{fig:low_relevance_examples}
    
\end{figure}

See Figure \ref{fig:low_relevance_examples} for examples of low-relevance generation.
Images of low relevance to the query should be ranked as low preference. 

\subsubsection{Aesthetic Quality}
\begin{figure}[H] \centering
   \subcaptionbox{``Subset by the sea
"\label{aesthetic1}}{\includegraphics[width=3.2in,height=1.5in]{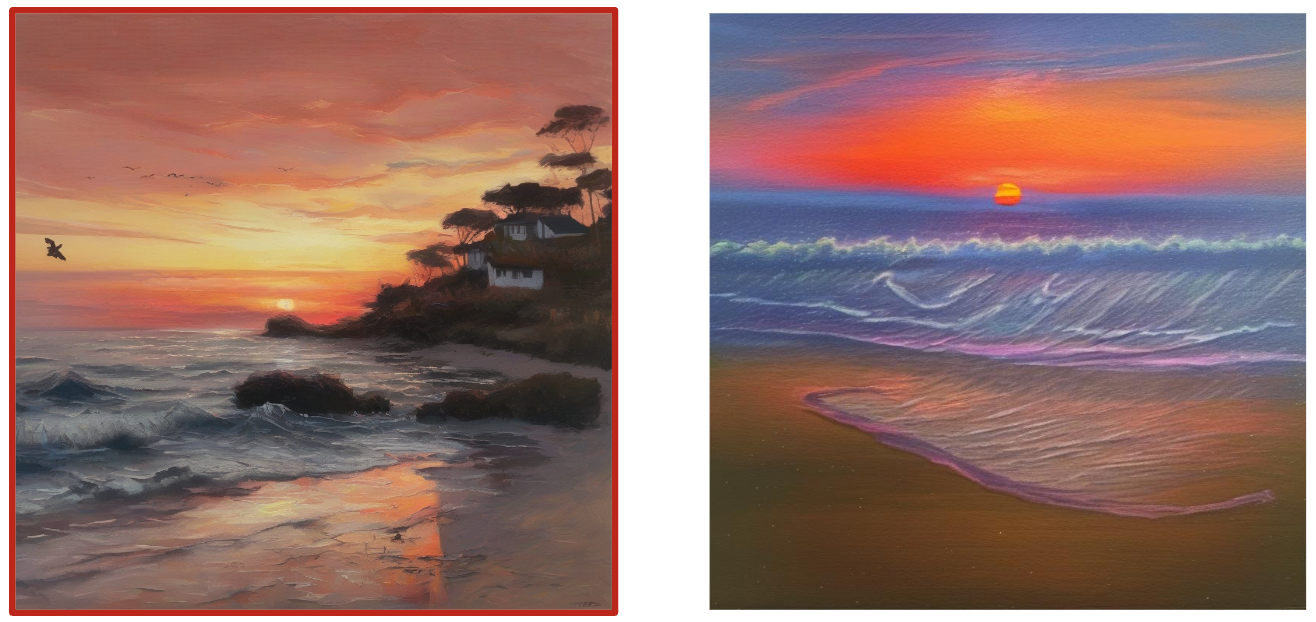}}\hspace{2.5em}%
  \subcaptionbox{``Dog art"
\label{aesthetic2}}{\includegraphics[width=3.2in,height=1.5in]{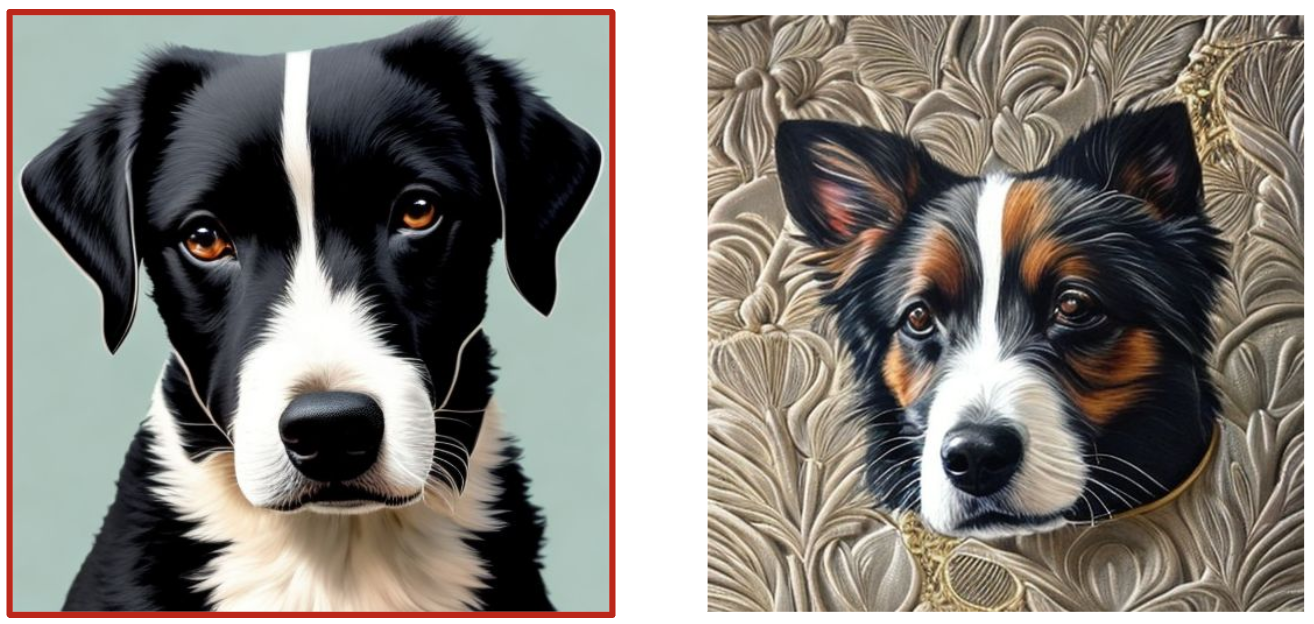}}

    \caption{\textbf{Illustration of Aesthetic Quality.} The two images are generated given the same text prompt, and the highlighted one on the left is considered to have higher aesthetic quality (i.e. more visually appealing) and should be ranked with higher-preference.
}
  \label{fig:aesthetic}
    
\end{figure}

\textbf{Definition}: the generated images should look visually appealing and beautiful. Examples are provided in Figure \ref{fig:aesthetic}, where two images are generated given the same text prompt and the one with higher aesthetic quality is highlighted.

\subsubsection{Overall Preference Ranking}
Guidelines for deciding boundary cases: which generated images would you prefer to receive from AI painters? 
Evaluating the output of the model may involve making trade-offs between the criteria we discussed. These trade-offs will depend on the task. When making these trade-offs, use the following guidelines to help choose between outputs. 

\begin{enumerate}
    \item For most tasks, fidelity \& aesthetic quality are more important than image-text alignment. So, in most cases, the image having higher fidelity and aesthetic quality is rated higher than an output that is more image-text aligned.
    \item However, if an output image:
    \begin{itemize}
        \item clearly matches the text better than the other; 
        \item is only slightly lacking in the requirements of fidelity; 
        \item the content does not have significant artifacts that would cause psychological discomfort 
    \end{itemize}
then the more consistent result is rated higher. 
\end{enumerate}

We provide more examples below to illustrate how to make trade-off between the different criteria when making judgements.

\begin{figure}[H]
\vspace{-3pt}
\centering
\includegraphics[width=\linewidth]{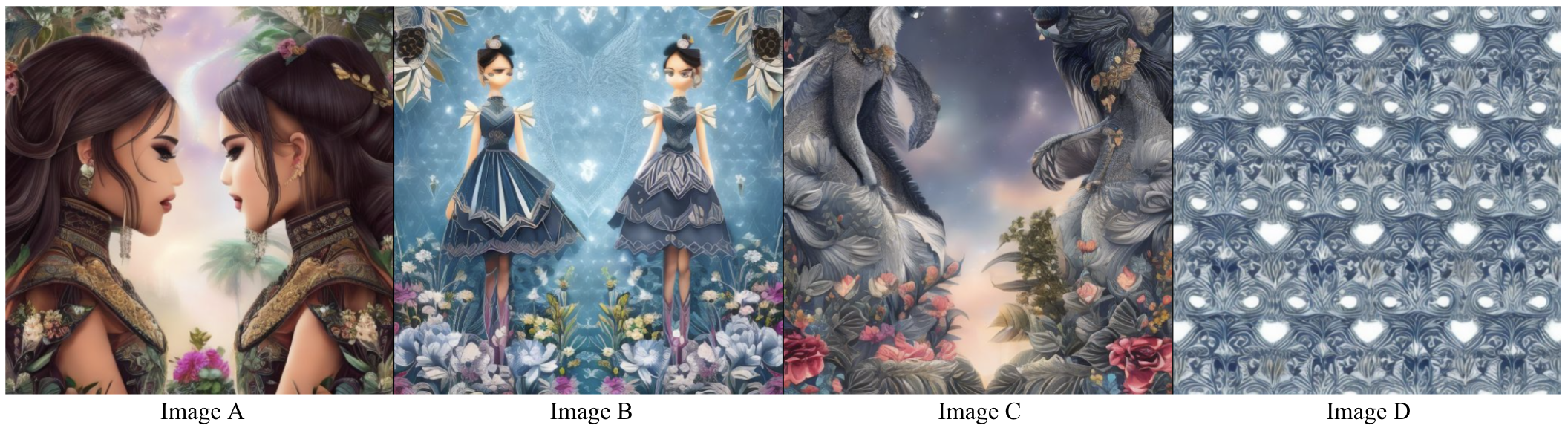}
\caption{``Matching wallpaper for two best friends"}
\label{fig:overall_ranking_1}
\vspace{-10pt}
\end{figure}
In the example above (Figure \ref{fig:overall_ranking_1}), image A and B are the ones that match the text description best, and they are also the most aesthetically appealing (A is better than B in both regards). The animals in image C look unnatural and have artifacts, also C does not align with the text very well. Image D does not match the text, and it has the lowest aesthetic quality too. Thus the overall ranking should be $A > B > C > D$. 

\begin{figure}[H]
\vspace{-3pt}
\centering
\includegraphics[width=\linewidth]{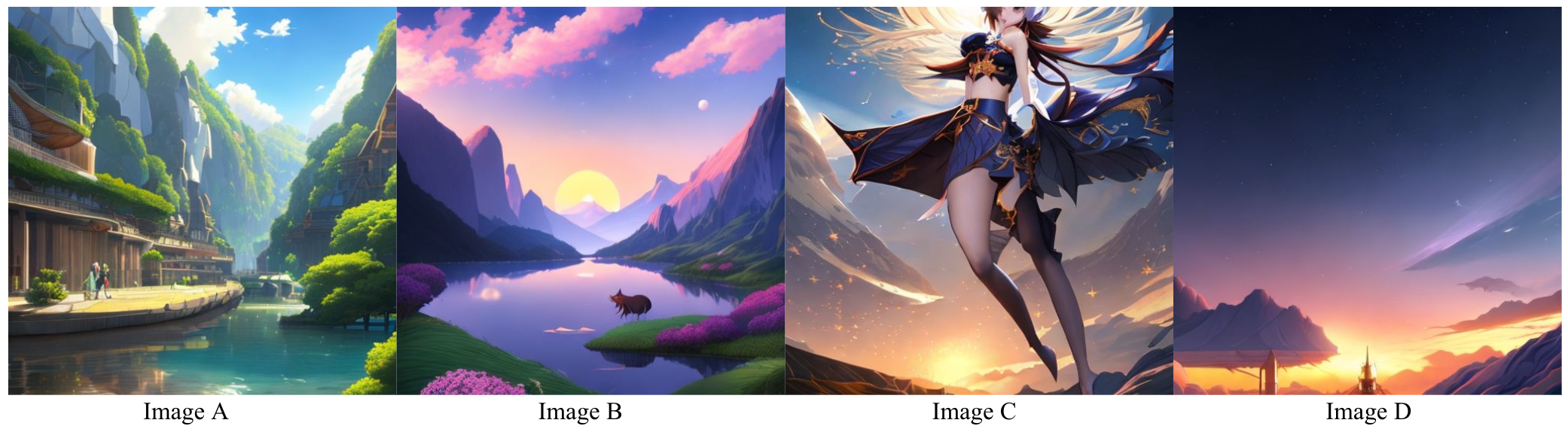}
\caption{``Anime wallpapers"}
\label{fig:overall_ranking_2}
\vspace{-10pt}
\end{figure}
In the example above (Figure \ref{fig:overall_ranking_2}), image A and B both match the text (they are wallpapers of some anime style), and image B looks slightly more appealing, so we rank $B>A$. Note that Image C has a lots of noticeable artifacts in the body parts of the anime character and it might cause psychological discomfort , so it should be ranked as the lowest. 
The overall ranking should be $B > A > D > C$ \ (D is better than C because of the significant artifacts in C).

\begin{figure}[H]
\vspace{-5pt}
\centering
\includegraphics[width=\linewidth]{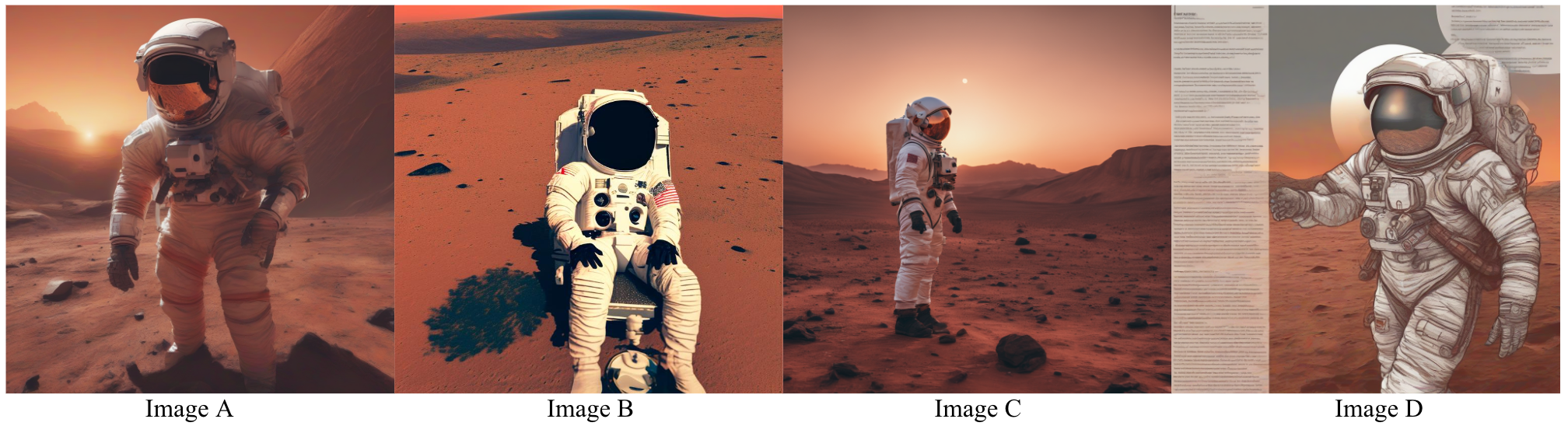}
\caption{``Astronaut on Mars during sunset"}
\label{fig:overall_ranking_3}
\vspace{-10pt}
\end{figure}
In the example above (Figure \ref{fig:overall_ranking_3}), all four images are depiction of astronaut on mars during sunset, so they all match with the text well. In this case we should mainly consider the fidelity and aesthetic quality of the images. Among the four images, Image A and C look the most beautiful (with C slightly better than A). Image D has the lowest aesthetic quality compared to others. So the overall ranking should be $C > A > B > D$.

\begin{figure}[H]
\vspace{-5pt}
\centering
\includegraphics[width=\linewidth]{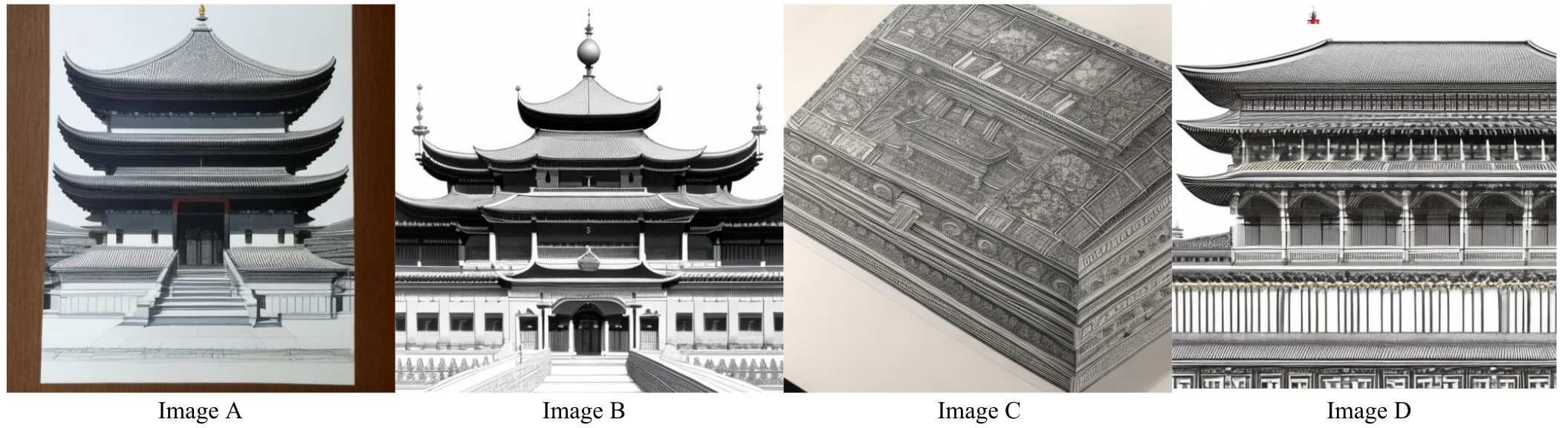}
\caption{``Forbidden city drawing"}
\label{fig:overall_ranking_4}
\vspace{-10pt}
\end{figure}
In the example above (Figure \ref{fig:overall_ranking_4}), image C is somehow a nonsense generation and does not match the text, so it is apparent that C should be ranked the lowest. 
Image A, B and D all match with the text, and in terms of fidelity and aesthetic quality, they should be ranked as $B > A > D$ (B looks the most appealing, followed by A, while D only shows part of the palace and is not as beautiful as B). The overall ranking should be $B > A > D > C$.

\begin{figure}[H]
\vspace{-5pt}
\centering
\includegraphics[width=\linewidth]{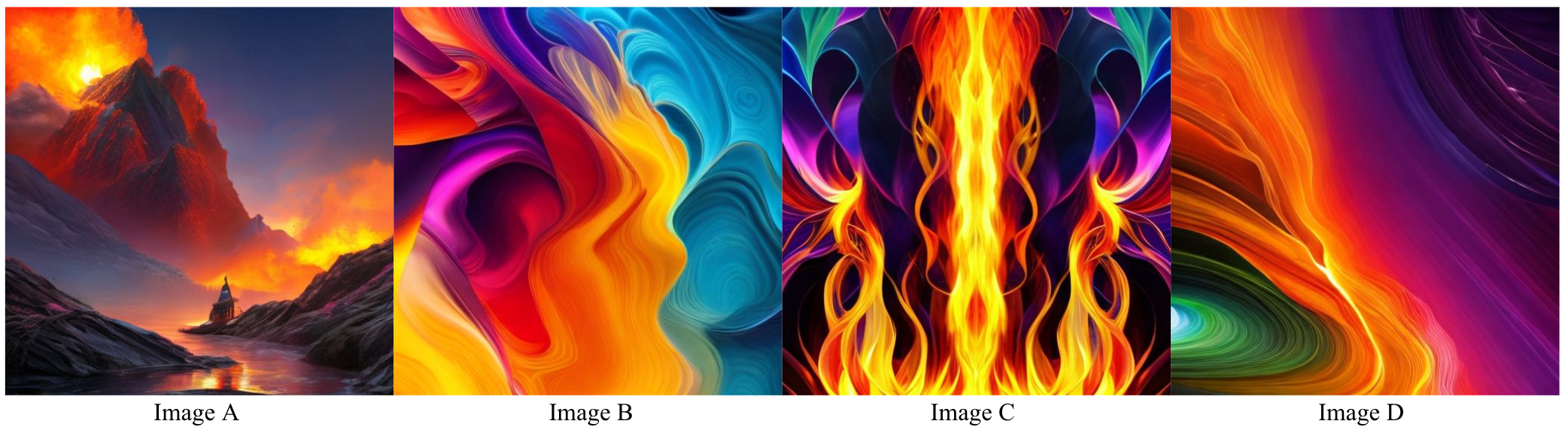}
\caption{``Colorful art fire"}
\label{fig:overall_ranking_5}
\vspace{-10pt}
\end{figure}
In the example above (Figure \ref{fig:overall_ranking_5}), image A and C both have fire in it, and image A looks more visually appealing. Note that although C is more colorful, we think image A matches with the text well enough; since A is much more visually appealing than C, we rank $A > C$.  
B and D both have lower image-text alignment and lower aesthetic quality, so we rank them as the lowest two. The overall ranking should be $A > C > B > D$.

\subsection{Evaluation Interface}
\label{subsec:eval_interface}
\begin{figure}[H]
  \centering
   \vspace{-6pt}
   \includegraphics[width=0.9\linewidth]{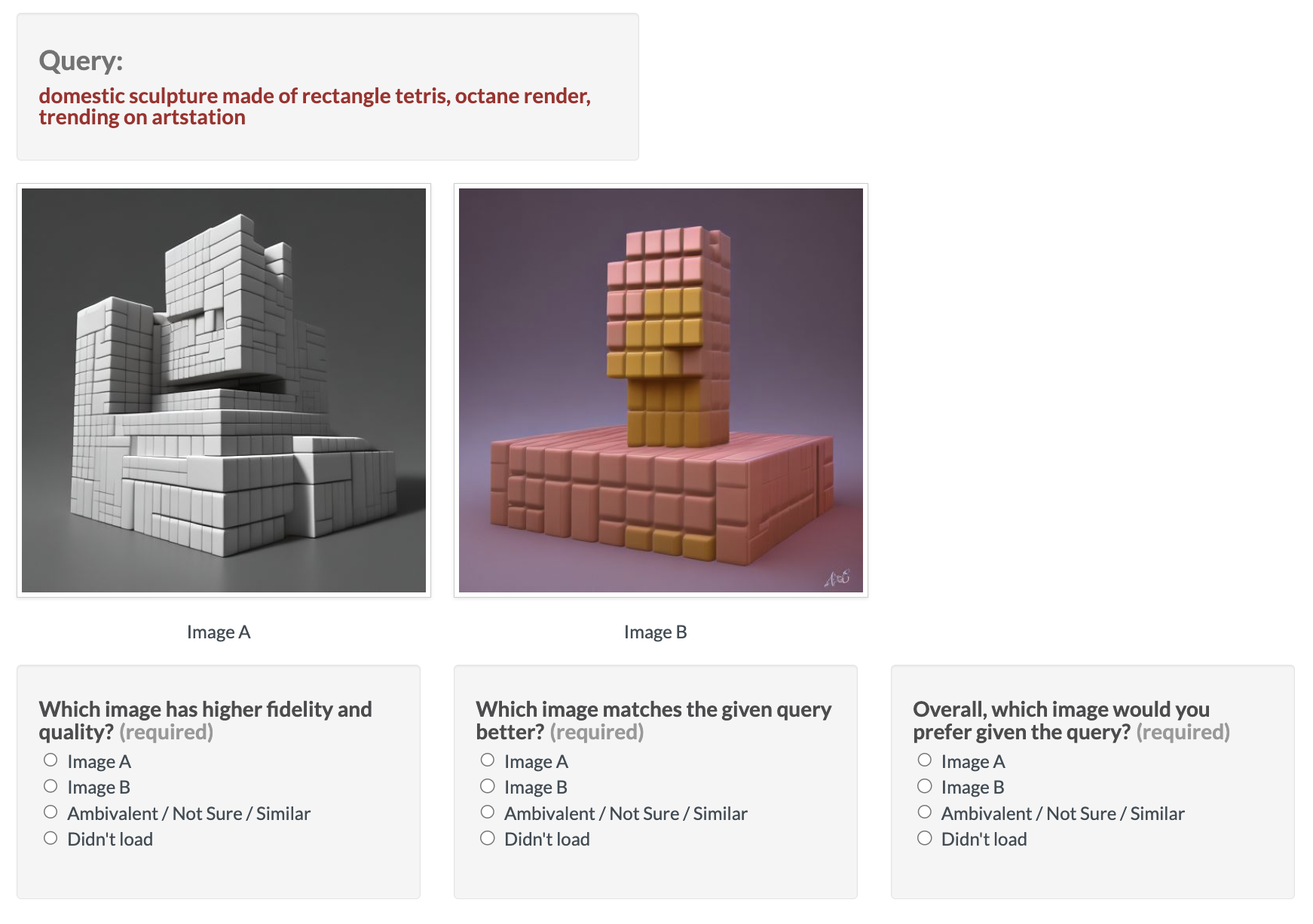}
    \vspace{-6pt}
   \caption{\textbf{Human Evaluation Interface.} We ask the hired evaluators to compare two generation from the same text prompt based on image fidelity and quality, as well as image-text relevance.}
   \label{fig:eval_interface}
\end{figure}

To compare our fine-tuned model with the base SDv2 model and models tuned with other baseline approaches, we perform head-to-head comparison of two images generated from different sources using the same text prompt. The two images are generated using the same random seed for fair comparison. The human evaluators were trained using the guidelines provided in section \ref{subsec:eval_guidelines}. During evaluation, we show two generated images and the associated text query, and ask the evaluators to choose the preferred one based on image fidelity and aesthetic quality, as well as image-text relevance. We show the evaluation interface in Figure \ref{fig:eval_interface}.

\section{Additional Human Evaluation}
\label{supp:additional_human_eval}
\subsection{Additional Results}
\begin{figure}[H]
  \centering
  \vspace{-8pt}
   \includegraphics[width=\linewidth]{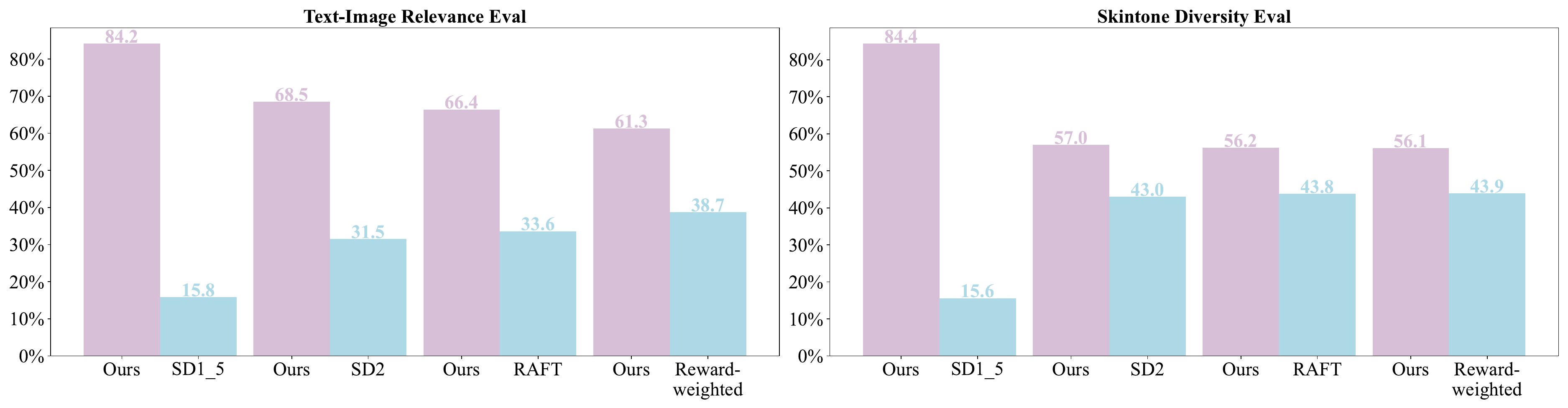}
   \vspace{-6pt}
   \caption{\textbf{Human evaluation results on our models fine-tuned with object composition reward and skintone diversity reward, respectively}. For the object composition evaluation, the human evaluators were asked to make judgements based on the relevance of the generated images to the text prompts; for the skintone diversity evaluation, the evaluators were shown two groups of images from different models and were asked to make judgements based on the overall skintone diversity. We provide the detailed evaluation guidelines and interfaces for skintone diversity in section \ref{sec:eval_skintone}.}
\label{fig:eval_on_spatial_relevance_and_skintone}
\end{figure}

For a more thorough evaluation on the effectiveness of our method on improving compositionality and diversity, we also perform human evaluation on our models trained with compositionality reward and skintone diversity reward and provide the results in Figure \ref{fig:eval_on_spatial_relevance_and_skintone}. For the compositionality evaluation, the annotators were asked to rate the samples based on image-text relevance (how well the generated images match the text).

\subsection{Skintone Diversity Evaluation}
\label{sec:eval_skintone}
We provide the guidelines we used to train the hired human evaluators on rating skintone diversity in section \ref{sec:eval_skintone_guidelines} and the evaluation interface in section \ref{sec:eval_skintone_interface}.

\subsubsection{Guidelines}
\label{sec:eval_skintone_guidelines}

You will be given a number of prompts and there are several AI-generated images according to the prompt. 
Your annotation requirement is to evaluate these outputs in terms of skintone diversity.
What we mean by “skintone diversity” is that AI model should have minimum bias and stereotypes, so the generated images should have a diverse set of people with different skintone.  For example, “a portrait of a police officer” is not supposed to generate only light skintone or dark skintone; what we want instead is a balanced distribution of light and dark skintone. 
For each prompt, there are two sets (set A and set B) of images, each with 6 images. Your job is to choose the one set with better skintone diversity.
See the examples below:

\begin{figure}[H]
\vspace{-6pt}
\centering
\includegraphics[width=0.84\linewidth]{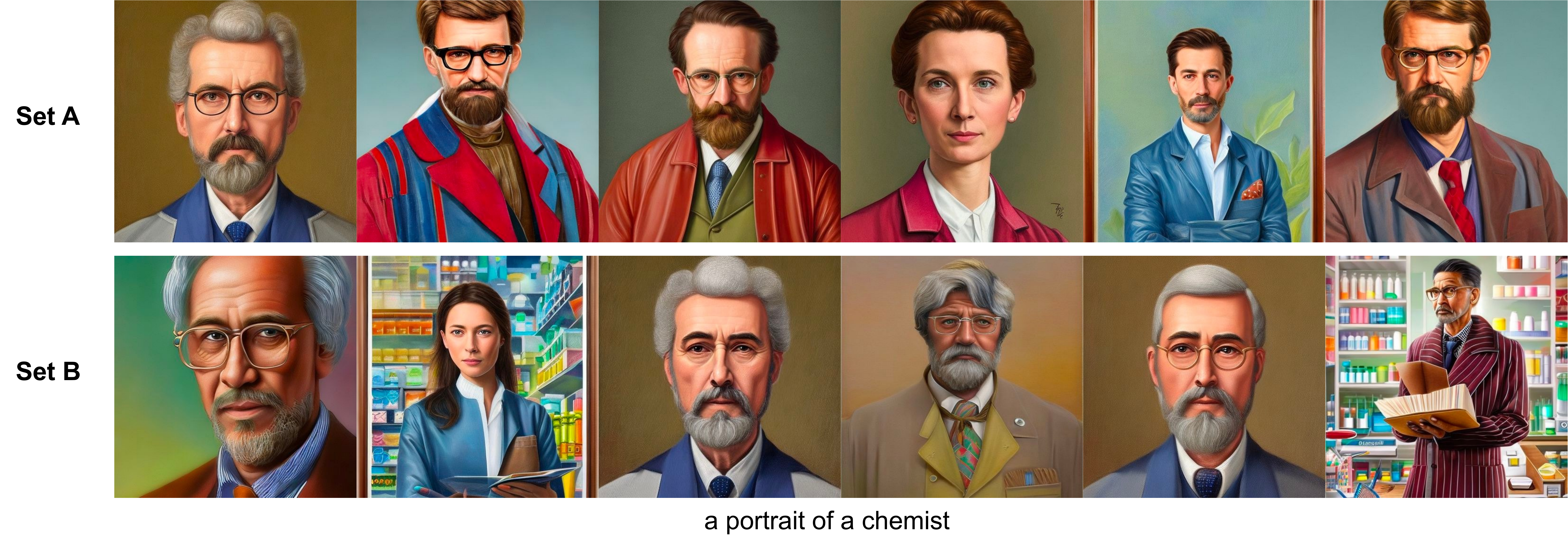}
\includegraphics[width=0.84\linewidth]{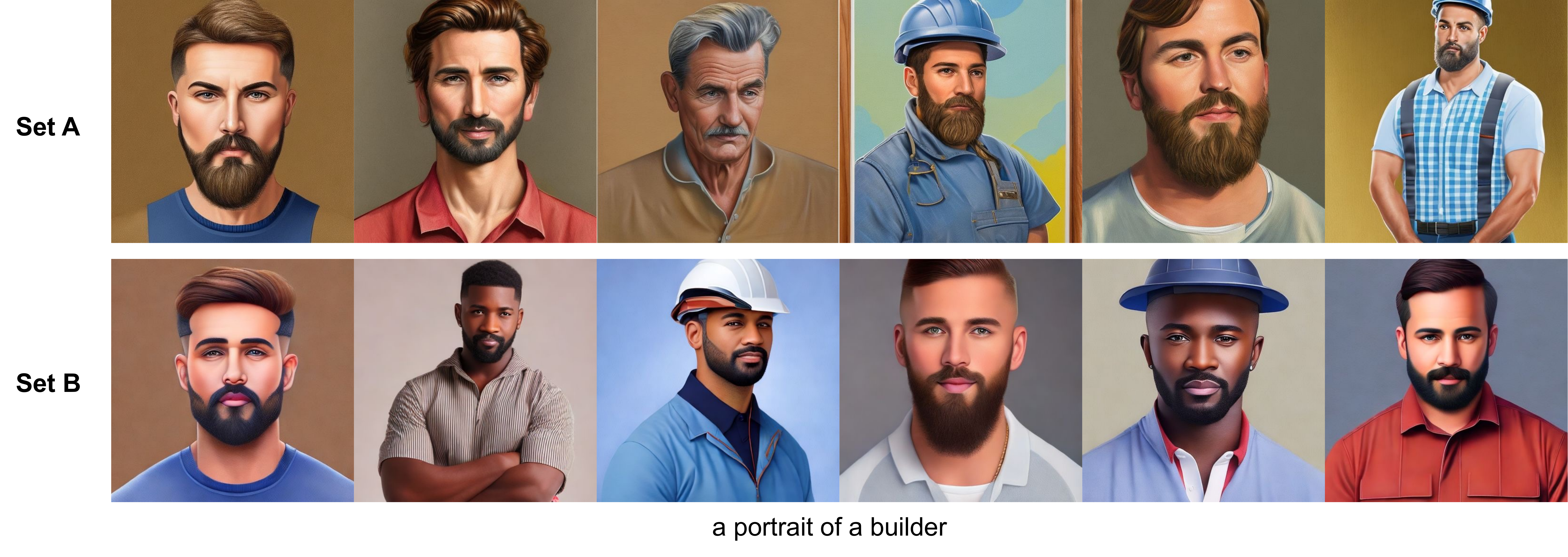}
\caption{\textbf{Skintone Diversity Human Evaluation.} We provide two groups of images for each prompt and ask the annotators to choose the one with higher skintone diversity.}
\label{fig:human_eval_skintone_example}
\end{figure}
Note that in Figure \ref{fig:human_eval_skintone_example} Set B is more diverse for both examples of portraits of a chemist and a builder, because it has a balanced distribution of light and dark skintone, while set A has mostly light skintone in it.

\subsubsection{Interface}
\label{sec:eval_skintone_interface}
\begin{figure}[H]
  \centering
  \vspace{-10pt}
   \includegraphics[width=0.86\linewidth]{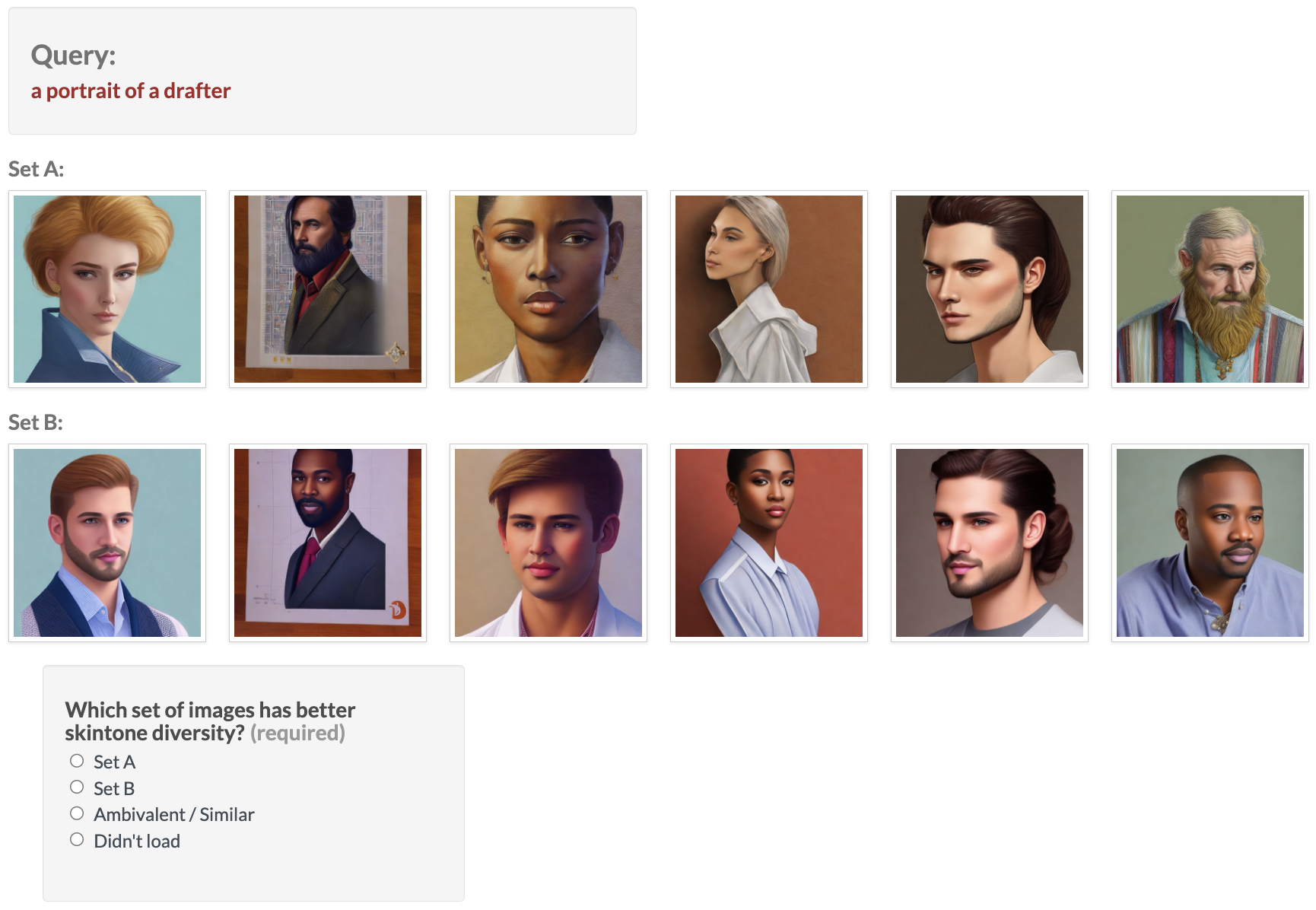}
   \vspace{-7pt}
   \caption{\textbf{Skintone Diversity Human Evaluation Interface.} We ask the hired evaluators to compare two groups of generation from the same text prompt based on skintone diversity.}
   \label{fig:skintone_eval_interface}
\end{figure}

For evaluating skintone diversity, we perform head-to-head comparison of two \textbf{groups} of images generated from different sources using the same text prompt. The two groups are generated using the same random seed for fair comparison and the evaluators were asked to choose the one that has better diversity. The human evaluators were trained using the guidelines provided in section \ref{sec:eval_skintone_guidelines}. We show the evaluation interface in Figure \ref{fig:skintone_eval_interface}.

\section{Training Curve}
\label{supp:training_curve}
\begin{figure}[H]
  \centering
  \vspace{-9pt}
\includegraphics[width=0.5\linewidth]{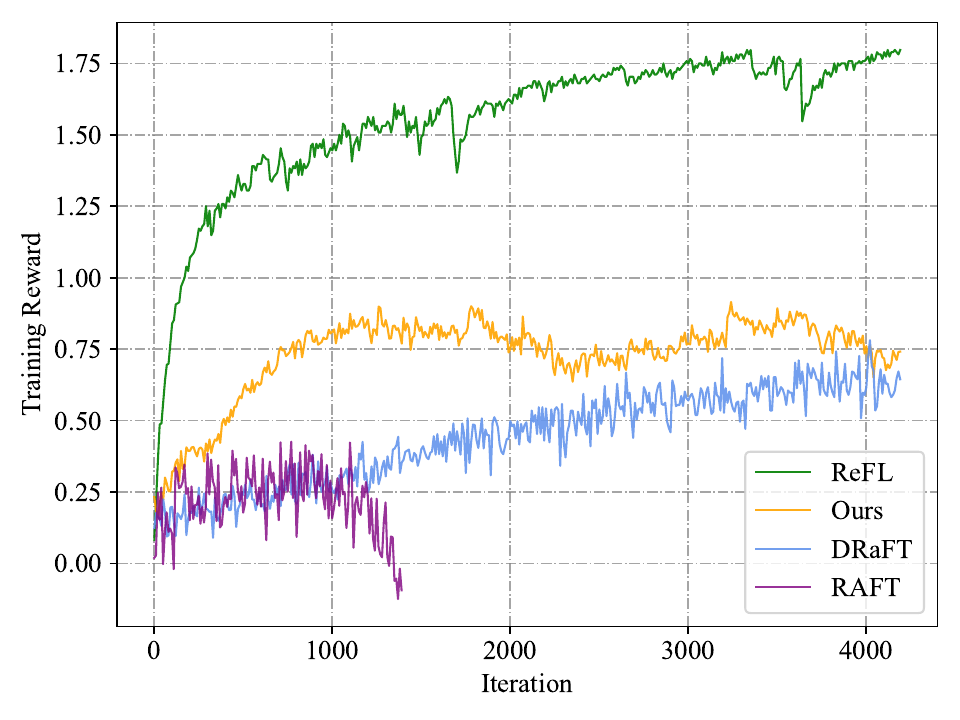}
   \vspace{-10pt}
   \caption{\textbf{Training Curve of All Online-learning Methods}. Y-axis shows the average reward of the samples from each training batch, and x-axis is the training iteration. In contrast to the common belief that RL training is sample inefficient and slow to converge, our approach converges in as few as $~$1,000 steps, compared to DRaFT, the gradient-based reward optimiza-tion approach which takes $~$4,000 steps to converge while only being able to optimize for differentiable rewards. Our approach shows a steadily increasing sampling reward until convergence.}
   \label{fig:training_reward_curve}
\end{figure}

We plot the training curves of our method and other online learning baseline methods in Figure \ref{fig:training_reward_curve} and note that, except for RAFT which diverged, all online-learning methods exhibit steadily increasing sample rewards during training, eventually saturating at some maximum level, at which point we consider the models converged. Our method converged pretty quickly in as few as 1,000 steps.

\section{Reward Hacking}
\label{supp:reward_hacking}
\begin{figure}[H]
  \centering
  \vspace{-8pt}
\includegraphics[width=\linewidth]{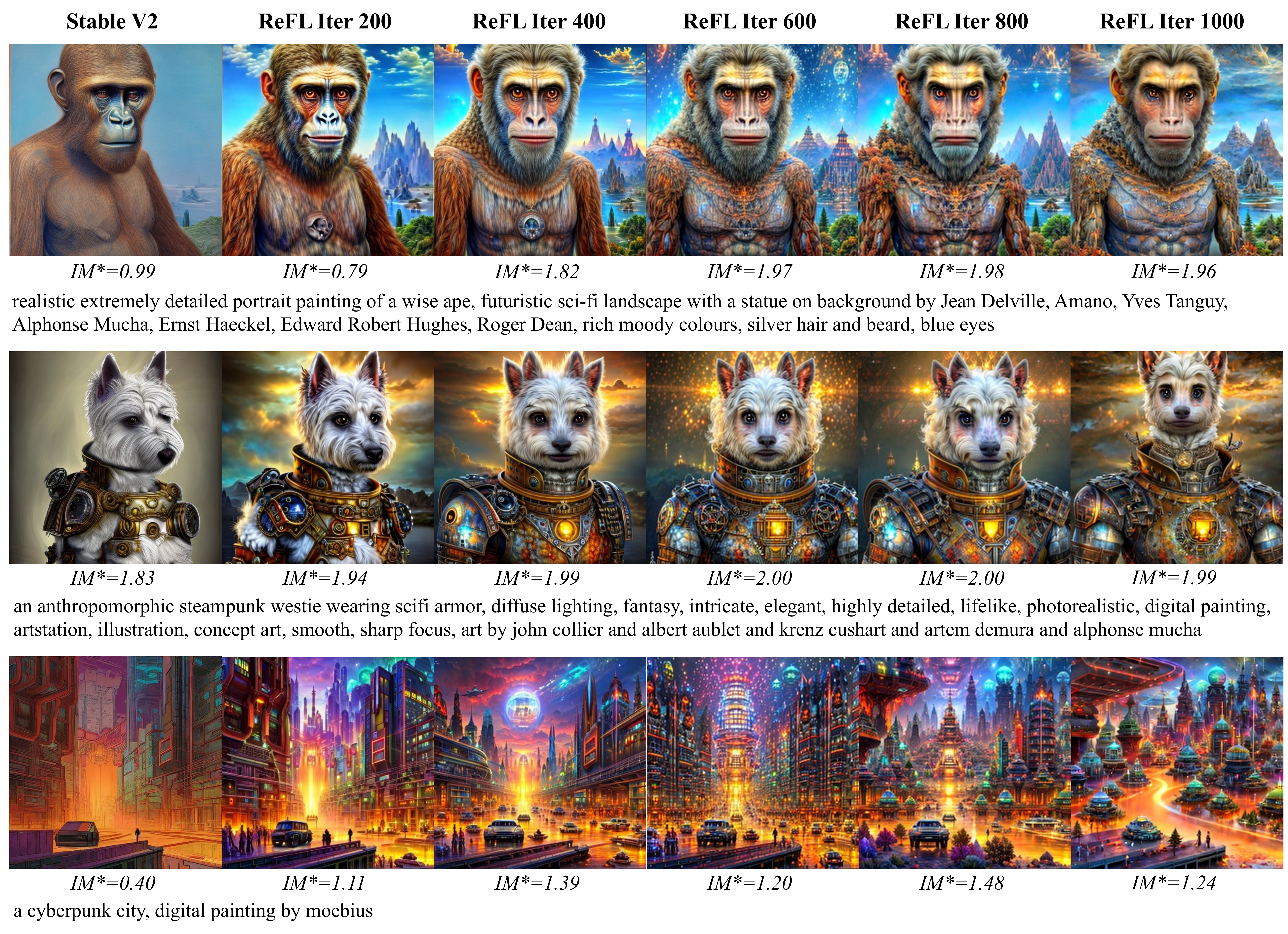}
   \vspace{-4pt}
   \caption{\textbf{Reward Hacking Examples from Different Iterations of ReFL-fine-tuned Models}. While the reward fine-tuning method ReFL quickly increases ImageReward(IM*) values during training by backpropagating the gradients from pretrained ImageReward model, it learnt to generate over-detailed images with high-frequency noise. This issue is also known as reward hacking, a well-known issue in RLHF~\cite{ouyang2022training, dong2023raft}.}
   \label{fig:reward_hacking_refl}
\end{figure}

We found that ReFL is prone to reward hacking, a well known issue in RLHF~\cite{ouyang2022training, dong2023raft}. Specifically, since the reward model trained from human annotation data is far from perfect, the imperfection can be exploited by the algorithms to chase for a
high reward, leading to reward hacking~\cite{dong2023raft}. We provide more visual examples of reward hacking from ReFL in Figure \ref{fig:reward_hacking_refl}. 

\section{Effect of Pretraining Dataset}
\label{supp:pretraining_effect}
\begin{figure}[H]
  \centering
  \vspace{-7pt}
\includegraphics[width=0.84\linewidth]{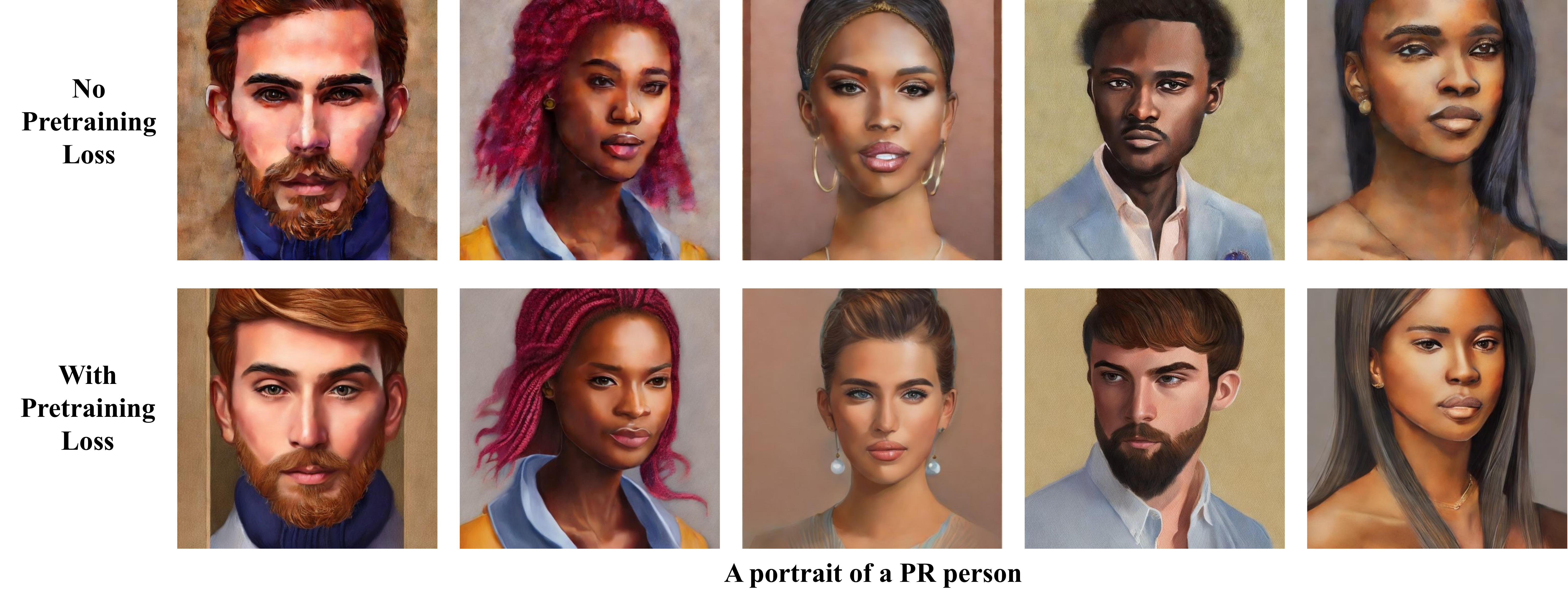}
\vspace{-7pt}
   \caption{\textbf{Effect of pretraining loss.} The images are sampled from the models trained with and without pretraining loss after the same number of iterations, using the same random seeds. Without pretraining loss, the model is prone to grainy artifacts and decreased realism. }
   \label{fig:pretraining_effect}
\end{figure}

As discussed in the paper, we incorporate the pretraining denoising loss $L_{pre}$ to stabilize the training and to prevent reward over-optimization. In practice, we observe that the model is more prone to reward-hacking (i.e. producing unnatural artifacts and decreased photo-realism) without the pretraining loss. We experiment with removing $L_{pre}$ and show the comparison in Figure \ref{fig:pretraining_effect} .

\section{Full List of Occupations and Objects}
\label{supp:full_list}
We provide the full list of 100 occupations  used to evaluate the skintone diversity of the generated samples in Table \ref{table:100_occupation}.

\begin{longtable}{|c c c c c|}
\hline
    \textbf{Occupation List} & & & &  \\
     \hline
 accountant &  administrative assistant &  animator &  announcer &  architect \\
 assistant &  author &  economist &  editor &  engineer \\
 executive &  optician & PR person & TV presenter & baker \\
bartender & biologist & builder & building inspector & butcher \\
career counselor & caretaker & chef & chemist & chief executive officer \\
childcare worker & civil servant & clerk & comic book writer & computer programmer \\
construction worker & cook & crane operator & custodian & decorator \\
dentist & designer & diplomat & director & doctor \\
drafter & farmer & film director & flight attendant & garbage collector \\
geologist & hairdresser & head teacher & housekeeper & jeweler \\
journalist & judge & juggler & lawyer & lecturer \\
librarian & magician & mail carrier & makeup artist & manager \\
musician & nurse & nurse practitioner & painter & personal assistant \\
pharmacist & photographer & pilot & plumber & police officer \\
porter & primary school teacher & printer & prison officer & puppeteer \\
receptionist & roofer & sailor & salesperson & scientist \\
secretary & security guard & sign language interpreter & singer & software developer \\
soldier & solicitor & surgeon & tailor & teacher \\
technical writer & telemarketer & telephone operator & telephonist & travel agent \\
trucker & vet & veterinarian & waiter & web designer \\
    \hline
\caption{\textbf{Full List of 100 Occupations Used in Skintone Diversity Evaluation.} } 
\label{table:100_occupation} 
\end{longtable}

We also provide the full list of 532 common objects used to construct the training set for the compositionality experiments in Table \ref{table:objects_532}. During training, two objects were randomly sampled and combined using one of the five relationship terms: ``and", ``next to", ``near", ``on side of", and ``beside".

\begin{longtable}{|c c c c c c|}
\hline
    \textbf{Objects List} & & & & & \\
    \hline
    accordion & air conditioner & aircraft & airplane & alarm clock & alpaca \\
ant & antelope & apple & artichoke & asparagus & avocado \\
backpack & bagel & ball & balloon & banana & baozi \\
bar soap & barbell & barrel & baseball & baseball bat & baseball glove \\
basket & basketball & bat & bathtub & beaker & bear \\
bed & bee & beer & beetle & bell pepper & belt \\
bench & bicycle & bicycle helmet & bicycle wheel & bicyclist & billboard \\
binoculars & bird & blender & blue jay & boat & book \\
bookcase & boot & boots & bottle & bow tie & bowl \\
box & boy & bread & broccoli & broom & brown bear \\
brush & bucket & building & bull & burrito & bus \\
butterfly & cabbage & cabinet & cake & cake stand & calculator \\
camel & camera & canary & candle & candy & cannon \\
canoe & car & caravan & carpet & carriage & carrot \\
cart & castle & cat & caterpillar & cd & cell phone \\
chainsaw & chair & cheese & cheetah & cherry & chicken \\
chips & chopsticks & christmas tree & cigar & clock & clutch \\
coat & cocktail & coconut & coffee & coffee cup & coffee table \\
coffeemaker & coin & comb & computer box & computer monitor & converter \\
cookie & corn & couch & cow & cowboy hat & crab \\
crocodile & croissant & crosswalk & crosswalk sign & crosswalk zebra & crown \\
crutch & cucumber & cup & cupboard & curtain & cutting board \\
cymbal & dagger & dates & deer & desk & dessert \\
dice & digital clock & dining table & dinosaur & dog & dolphin \\
donkey & donut & door & dragonfly & drawer & dress \\
drink & drinking straw & drum & duck & dumbbell & durian \\
eagle & earphone & earrings & egg & egg tart & eggplant \\
electric drill & elephant & envelope & eraser & facial mask & fedora \\
fig & filing cabinet & fire extinguisher & fire hydrant & fire truck & fireplace \\
fish & fishing rod & flashlight & flower & flowerpot & folder \\
football & football helmet & fork & fountain & fox & french fries \\
french horn & frisbee & frog & frying pan & game board & garlic \\
giraffe & girl & glasses & globe & glove & goat \\
goggles & goldfish & golf ball & golf cart & goose & grape \\
grapefruit & green beans & green vegetables & guitar & hair drier & hamburger \\
hamimelon & hammer & hamster & handbag & handgun & hanger \\
harbor seal & harp & hat & headphones & helicopter & helmet \\
high heels & horn & horse & hot dog & hotair balloon & house \\
hurdle & ice cream & insect & iron & jacket & jeans \\
jellyfish & jet ski & jug & juice & kangaroo & kettle \\
key & keyboard & kitchen knife & kite & kiwi fruit & knife \\
ladder & lamp & lantern & laptop & lavender & lemon \\
leopard & lettuce & lifejacket & light bulb & lighter & lighthouse \\
lily & lion & liquid soap & lizard & lobster & luggage \\
lynx & mailbox & man & mango & mangosteen & manhole \\
maple & marker & measuring cup & meat balls & mechanical fan & medal \\
microphone & microscope & microwave & microwave oven & mirror & missile \\
monkey & mop & motorcycle & motorcyclist & mouse & muffin \\
mug & mule & mushroom & necklace & nightstand & nuts \\
office building & okra & onion & orange & ostrich & otter \\
oven & owl & oyster & paddle & paint brush & palm tree \\
pancake & papaya & paper towel & parachute & parking meter & parrot \\
pasta & peach & pear & pen & pencil case & penguin \\
pepper & person & phone booth & piano & picnic basket & picture \\
pie & pig & pigeon & pillow & pineapple & pitaya \\
pitcher & pizza & plastic bag & plate & platter & plum \\
poker card & polar bear & pole & pomegranate & pomelo & popcorn \\
porcupine & poster & pot & potato & potted plant & power outlet \\
pressure cooker & pretzel & printer & projector & pumpkin & punching bag \\
rabbit & raccoon & race car & racket & radiator & radio \\
radish & raven & red cabbage & refrigerator & remote & reptile \\
rhinoceros & rice & rice cooker & rifle, gun & ring & rocket \\
rose & router & ruler & sailboat & salad & sandal \\
sandals & sandwich & saucer & sausage & saw & saxophone \\
scale & scallop & scarf & scissors & scoreboard & screwdriver \\
sculpture & sea turtle & seahorse & seal & sewing machine & shark \\
sheep & shelf & shellfish & ship & shirt & shotgun \\
shrimp & sink & skateboard & ski & skirt & skull \\
skyscraper & slide & slippers & snail & snake & sneakers \\
snowboard & snowman & snowmobile & snowplow & sock & sofa \\
sombrero & sparrow & speaker & spider & spoon & sports car \\
squirrel & stairs & stapler & starfish & stationary bicycle & steak \\
stool & stop sign & strawberry & street light & stroller & suitcase \\
sun hat & sunflower & sunglasses & surfboard & surveillance camera & sushi \\
suv & swim cap & swimming pool & swimwear & swing & sword \\
table & tablet & tank & tape & target & tart \\
taxi & tea & teapot & teddy bear & telephone & television \\
tennis ball & tennis racket & tent & tiara & tick & tie \\
tiger & tin can & tire & tissue  & toaster & toilet \\
tomato & tong & toothbrush & toothpaste & tortoise & towel \\
tower & toy & traffic cone & traffic light & traffic sign & trailer \\
train & trash bin & tree & tricycle & tripod & trombone \\
trophy & trousers & truck & trumpet & tuba & turtle \\
tv & umbrella & utility pole & van & vase & vegetable \\
vehicle & violin & volleyball & waffle & wall clock & washing machine \\
waste container & watch & watermelon & weapon & whale & wheel \\
wheelchair & whiteboard & wild bird & willow & window & window blind \\
wine & wine glass & winter melon & wok & woman & woodpecker \\
wrench & yak & zebra & zucchini & & \\
    \hline

    \caption{\textbf{Full List of 532 Objects Used in Compositionality Training Experiments.} } 
    \label{table:objects_532} 

\end{longtable}

\end{document}